%% file: neurips_2024.tex
\documentclass{article}

\usepackage[final]{neurips_2024}
\usepackage{multirow}
\usepackage[utf8]{inputenc} % allow utf-8 input
\usepackage[T1]{fontenc}    % use 8-bit T1 fonts
\usepackage{hyperref}       % hyperlinks
\usepackage{url}            % simple URL typesetting
\usepackage{booktabs}       % professional-quality tables
\usepackage{amsfonts}       % blackboard math symbols
\usepackage{nicefrac}       % compact symbols for 1/2, etc.
\usepackage{microtype}      % microtypography
\usepackage{xcolor}         % colors
\usepackage{amsmath}        % amsmath
\usepackage{algorithm}       % For writing algorithms
\usepackage{algpseudocode} % For pseudocode
\usepackage{array}
\usepackage{graphicx}
\usepackage{wrapfig}
\usepackage{caption}
\usepackage{enumitem}
\input{macro}

\title{Diffusion-based Layer-wise Semantic Reconstruction for Unsupervised Out-of-Distribution Detection}

\author{
\textbf{Ying Yang}$^{1\mathrm{\ast}}$, 
\textbf{De Cheng}$^{1\mathrm{\ast} \dagger}$, 
\textbf{Chaowei Fang}$^{1\mathrm{\ast} \dagger}$, 
\textbf{Yubiao Wang}$^{1}$ \\
\textbf{Changzhe Jiao}$^{1}$, 
\textbf{Lechao Cheng}$^{2}$,
\textbf{Nannan Wang}$^{1}$, 
\textbf{Xinbo Gao}$^{3}$ \\
$^{1}$Xidian University \\
$^{2}$Hefei University of Technology \\
$^{3}$Chongqing University of Posts and Telecommunications\\
\texttt{\{yycfq, ybwang\_3\}@stu.xidian.edu.cn} \\
\texttt{chenglc@hfut.edu.cn}, \texttt{gaoxb@cqupt.edu.cn}\\
\texttt{\{dcheng, cwfang, cjiao, nnwang\}@xidian.edu.cn} 
}

\begin{document}

\maketitle
\makeatletter
\renewcommand\@makefnmark{}
\makeatother
% \footnotetext{$^{\ast}$ Equation Contribution.}
\footnotetext{$^{\ast}$ Equation Contribution. $^{\dagger}$ Corresponding authors: De Cheng and Chaowei Fang.}

\begin{abstract}
Unsupervised out-of-distribution (OOD) detection aims to identify out-of-domain data by learning only from unlabeled In-Distribution (ID) training samples, which is crucial for developing a safe real-world machine learning system. Current reconstruction-based method provides a good alternative approach, by measuring the reconstruction error between the input and its corresponding generative counterpart in the pixel/feature space. However, such generative methods face the key dilemma, $i.e.$, \emph{improving the reconstruction power of the generative model, while keeping compact representation of the ID data.} To address this issue, we propose the diffusion-based layer-wise semantic reconstruction approach for unsupervised OOD detection. The innovation of our approach is that we leverage the diffusion model's intrinsic data reconstruction ability to distinguish ID samples from OOD samples in the latent feature space. Moreover, to set up a comprehensive and discriminative feature representation, we devise a multi-layer semantic feature extraction strategy. Through distorting the extracted features with Gaussian noises and applying the diffusion model for feature reconstruction, the separation of ID and OOD samples is implemented according to the reconstruction errors. Extensive experimental results on multiple benchmarks built upon various datasets demonstrate that our method achieves state-of-the-art performance in terms of detection accuracy and speed. Code is available at \href{https://github.com/xbyym/DLSR}{https://github.com/xbyym/DLSR}.
\end{abstract}
\section{Introduction}
Unsupervised Out-of-Distribution (OOD) detection aims to identify whether a data point belongs to the In-Distribution (ID) or OOD dataset, by learning only from unlabeled in-distribution training samples. 
OOD detection plays a vital role in developing a safe real-world machine learning system, which ensures that the model is only performed on data drawn from the same distribution as its training data. 
If the test data does not follow the training distribution, the model could unintentionally produce nonsensical predictions, resulting in some misleading conclusions. Naturally, OOD detection is one of the key techniques for ensuring the model's robustness and safety. 

Existing research studies the OOD detection mainly under two settings, $i.e.$, supervised and unsupervised. 

The supervised OOD detection methods usually deem this task as a binary classification problem, which relies on training with data labeled as OOD from disjoint categories or adversaries \citep{hendrycks2018deep}, \citep{ming2022poem}. However, in many practical applications, it is almost impossible to access representative OOD samples, as the OOD data usually can be highly diverse and unpredictable.
Therefore, we prefer to study the more challenging while practical unsupervised OOD detection problem. We will build an OOD detector trained solely on unlabeled ID data, as large amounts of unlabeled data are readily available and widely utilized due to their ease of acquisition.

Current reconstruction-based methods provide a good alternative approach for OOD detection, by measuring the reconstruction error between the input and its corresponding generative counterpart in the pixel/feature space. Obviously, the generative models and metric learning evaluation strategies are the main research directions. However, such methods of the generative models always face the following key dilemma: The projected in-distribution latent feature space should be compressed sufficiently to capture the exclusive characteristics of ID images, while it should also provide sufficient reconstruction power for the large-scale ID images of various categories. Existing generative-based methods ($e.g.$, auto-encoder (AE), variational AE \citep{kingma2013auto} and Generative Adversarial Network(GAN)) \citep{NIPS2014_5ca3e9b1}, can not always fulfill these two requirements simultaneously, and a good balance between them is required. Besides, recent OOD detection methods based on diffusion models such as  \citep{graham2023denoising}, \citep{gao2023diffguard} and \citep{liu2023unsupervised} often involve the pixel-level reconstruction of distorted images, which consume much training/inference time and computation resources.

To address the above-mentioned issues, and inspired by the latent space noise addition mechanism in Latent Diffusion Models (LDM)~\cite{rombach2022high}, we propose the diffusion-based layer-wise semantic reconstruction approach for unsupervised OOD detection. Specifically, the proposed method makes full use of the diffusion model's intrinsic data reconstruction ability, to distinguish in-distribution samples from OOD samples in the latent feature space. In the diffusion denoising probabilistic models (DDPM) \citep{ho2020denoising}, the model is trained to incrementally remove noise from the noised inputs of different levels. Clearly, we can see that, instead of faithfully reconstructing inputs from the distribution it was trained on as previous VAE~\cite{kingma2013auto} or GAN~\cite{NIPS2014_5ca3e9b1}, the diffusion-based model shows more powerful reconstruction capabilities. Practically, our model involves reconstructing an input image feature from multiple values of the time step, this allows a single trained model to handle large amount of noise applied to the input, obviating the need for any dataset-specific fine-tuning. 

Moreover, to set up a comprehensive and discriminative feature representation, we devise a multi-layer semantic feature extraction strategy. Performing feature reconstruction on top of the multi-layer semantic features, encourages to restrict the in-distribution latent features distributed more compactly within a certain space, so as to better rebuild in-distribution samples while not reconstructing OOD comparatively. Overall, by distorting the extracted multi-layer features with Gaussian noises and applying the diffusion model for feature reconstruction, the separation of ID and OOD samples is implemented according to the reconstruction errors. Note that, the proposed Latent Feature Diffusion Network (LFDN) 
is built on top of the feature level instead of the traditional pixel level, which could significantly improve the computation efficiency and achieve effective OOD detection. The other potential strength of such a strategy is that it avoids the reconstruction of minor characteristics unrelated to image understanding. 
In summary, the contributions of this work are as follows:

\begin{itemize}
    \item We propose a diffusion-based layer-wise semantic reconstruction framework to tackle OOD detection, based on multi-layer semantic feature distortion and reconstruction. Meanwhile, We are the first to successfully incorporate generative modeling of features within the framework of OOD detection in image classification tasks.
    \item The layer-wise semantic feature reconstruction encourages restricting the in-distribution latent features to be more compactly distributed within a certain space, enabling better reconstruction of ID samples while limiting the reconstruction of OOD samples.
    \item Extensive experiments on multiple benchmarks across various datasets show that our method achieves state-of-the-art detection accuracy and speed.
\end{itemize}

\vspace{-4mm}
\section{Related Work}

\vspace{-2mm}
Existing researches study the OOD detection mainly under two settings: supervised and unsupervised.
The Supervised method is generally based on classification. The method usually uses the maximum softmax probability \citep{hendrycks2016baseline} from the final fully connected (FC) layer as the score to judge the ID sample. But the classification-based OOD detection methods often encounter issues with assigning high softmax probability to OOD  samples. Recent works \citep{liu2020energy}, \citep{sun2022dice}, \citep{djurisic2022extremely}, \citep{zhao2024towards}, attempt to alleviate this issue.
The unsupervised OOD detection can be roughly categorized as the distance-based metric evaluation and the generative-based reconstruction methods.

Distance-based methods assume that OOD data lies far from ID class centroids. \citep{ren2021simple} improved OOD detection by separating image foregrounds from backgrounds and computing the Mahalanobis distance for each, then combining them. \citep{sun2022out} used a non-parametric nearest neighbor distance for OOD detection. \citep{techapanurak2020hyperparameter} and \citep{chen2020boundary} used cosine similarity to measure distances between test data features of in-distribution data to identify OOD data. \citep{huang2020feature} applied Euclidean distance, while \citep{gomes2022igeood} used Geodesic distance for OOD detection. These methods often fail to capture sample distribution accurately.

Among the generative-based methods, the Likelihood-based methods can be traced back to as early as \citep{bishop1994novelty}. This method assumes that the generative model assigns high likelihood to ID data, while the likelihood for OOD data tends to be lower. Recently, several deep generative models have supported the computation of likelihood, such as VAE \citep{kingma2013auto}, PixelCNN++ \citep{salimans2017pixelcnn++}, and Glow \citep{kingma2018glow}. However, some studies (\citep{nalisnick2018deep}; \citep{choi2018waic}; \citep{kirichenko2020normalizing}) have found that probabilistic generative models might also assign high likelihood to OOD data.

A series of studies have attempted to mitigate this issue. \citep{serra2019} explored the relationship between image complexity and likelihood values, which adjusted likelihoods based on the size of image compression. \citep{ren2019} enhanced OOD detection by comparing likelihood values derived from different models. Another closely related approach highlights that these indicators are not well suited for VAEs. \citep{xiao2020likelihood} proposed a specialized metric known as likelihood regret for OOD detection in VAEs. \citep{cai2023out} suggested to leverage the high-frequency information of images to improve the model's ability to recognize OOD data. Additionally, a range of studies~\citep{nalisnick2019detecting}, \citep{wang2020further}, \citep{bergamin2022model}, \citep{osada2023out}, have proposed typicality tests, estimating layer activation distributions and other statistical measures on training data, which are then evaluated through hypothesis testing or density estimation.

Another type of OOD detection methods leverage the idea that generative networks produce different reconstruction errors for ID and OOD data. Some methods such as \citep{sakurada2014anomaly}, \citep{zong2018deep}, and \citep{zhou2017anomaly}, used auto-encoders to analyze reconstruction errors. GAN-based methods \citep{schlegl2017unsupervised}, \citep{zenati2018efficient}, and \citep{madzia2022progressive} utilized reconstruction errors and discriminator confidence to detect anomalies. Recent works \citep{graham2023denoising}, \citep{gao2023diffguard}, and \citep{liu2023unsupervised} applied diffusion models to model the pixel-level distribution of images, using errors from multiple reconstructions for OOD detection. Different from previous methods, we propose to leverage diffusion models to perform multi-layer semantic reconstruction in the latent feature space, 
not only for their stability in generation but also for significantly reducing training and inference time costs.

% Distance-based methods assume that OOD data lies far from ID class centroids. \citep{ren2021simple} improved OOD detection by separating image foregrounds and backgrounds and computing the Mahalanobis distance for each, then combining them. \citep{sun2022out} used a non-parametric nearest neighbor distance for OOD detection. \citep{techapanurak2020hyperparameter} and \citep{chen2020boundary} used cosine similarity to measure distances between test data features and in-distribution data to identify OOD data. \citep{huang2020feature} applied Euclidean distance, while \citep{gomes2022igeood} used Geodesic distance for OOD detection. These methods often fail to capture sample distribution accurately.

 % \note{\However, leveraging diffusion models for modeling multi-dimensional feature spaces is both important and reasonable, not only for their stability in generation but also for significantly reducing training and inference time costs.}

% \note{
% 为了增大不同类型变量之间的辨识度，根据以下原则对变量进行命名：

%  - 集合： 大写黑板加粗正体，e.g., $\mathbb X$

%  - 2维以上张量: 大写加粗正体，e.g., $\mathbf X$

%  - 1维向量： 小写加粗正体，e.g., $\mathbf x$

%  - 标量：不加粗斜体，e.g., $x$

%  - 函数：不加粗正体或者mathcal，e.g., $\textrm{ABC}(x)$, $\mathcal{F}(x)$

%  - 上标为数字时，添加括号避免和指数运算冲突, e.g., $x^{(1)}$

%  - 关键变量首次出现或未按照以上规则命名时，标明其维度, e.g., $\mathbf X \in \mathbb R^{a\times b\times c}$
%  }

\begin{figure}
    \centering
    \includegraphics[width=1\linewidth]{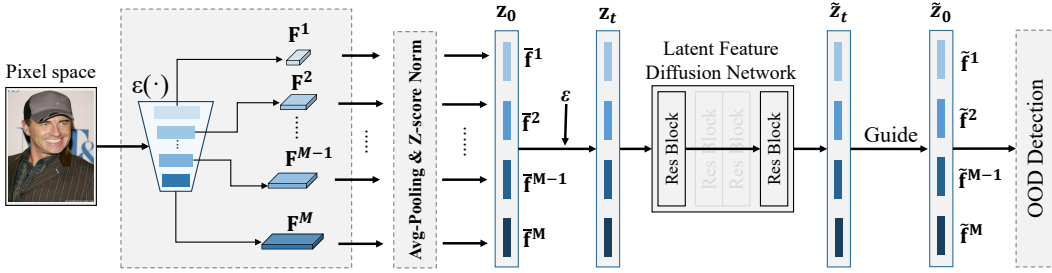}
    \caption{Overview of proposed diffusion-based layer-wise semantic reconstruction framework for unsupervised OOD detection. It includes multi-layer semantic feature extraction, Diffusion-based Feature Distortion and Reconstruction, and OOD detection head modules.}
    \label{fig:1}
\end{figure}
\vspace{-4mm}
\section{Method} 
\label{method}
Unsupervised OOD detection leverages intrinsic information from an unlabeled ID dataset \(\mathbb D\) to train a detector. Suppose $\mathbb D$ contains $N$ images, namely $\mathbb D=\{\textbf{x}_i\}_{i=1}^N$, where $\textbf{x}_i$ denotes the $i$-th image.
The target is to learn an OOD detector denoted as $\mathcal S(\cdot)$, which can effectively evaluate an OOD score for each input image.
The judgment of whether the input image belongs to ID or OOD is implemented by thresholding the OOD score.
For example, given a testing image $\textbf{x}$, it is recognized as an ID sample if the OOD score $\mathcal S(\textbf{x})$ is lower than the pre-defined threshold $\lambda$; otherwise, it is recognized as an OOD sample.

% \iffalse
% Denote the input image as $x$. The 

% This detector assesses new data by evaluating their OOD scores to determine if they belong to the distribution of \(D\). Higher scores are indicative of data more likely to be OOD. Thus, this can be defined as a binary classification problem facilitated by a thresholding mechanism:
% \[
% G_{\lambda}(x) = 
% \begin{cases} 
% \text{ID} & \text{if } S(x) \leq \lambda, \\
% \text{OOD} & \text{if } S(x) > \lambda,
% \end{cases}
% \]
% where data with lower scores \(S(x)\) are classified as ID, and those with higher scores as OOD. The threshold \(\lambda\) is typically set to correctly classify a high percentage, e.g., 95\%, of the ID data. Details on the computation of the OOD score are provided in Section 3.3.
% \fi
% To promote the application of diffusion model in OOD detection and reduce its training and inference cost, we propose a framework for modeling multi-layer low-dimensional features, which consists of two stages: semantic feature fusion and compression stage, and potential spatial diffusion stage.
In this paper, we propose a diffusion-based layer-wise semantic reconstruction framework to accomplish the OOD detection task. 
Specifically, as illustrated in Figure~\ref{fig:1}, the proposed framework consists of the following three components: the  multi-layer semantic feature extraction module, the latent feature diffusion stage, and the OOD  detection head. 

\subsection{ Multi-layer Semantic Feature Extraction}

The proposed semantic reconstruction-based method achieves OOD detection by measuring the reconstruction error between the input and its generative counterpart in the feature space. Specifically, we devise a multi-layer semantic feature extraction strategy, to set up a comprehensive and discriminative feature representation for each input image. Such multi-layer features could better rebuild the samples and encourage the ID semantic features distributed more compactly within a certain space from different semantic layers. %while not reconstructing 

Specifically, given an image $\textbf{x} \in \mathbb{R}^{3\times w \times h}$ with $w$ and $h$ being the width and height of the input image, passing through an image encoder $\mathcal{E}(\cdot)$, ($e.g.$, EfficientNet~\citep{tan2019efficientnet}), we can extract its feature maps from different layers ($i.e.$, low-level to high-level semantic blocks). The multi-layer intermediate feature map from the $m$-th block can be defined as $\mathbf{F}^m \in \mathbb{R}^{c_m \times w_m \times h_m}, m\in \{1,...,M\}$, where $c_m$, $w_m$ and $h_m$ are the number of channels, width and height of the feature map $\mathbf{F}^m$, and $M$ is the total number of intermediate feature maps. Then, each feature map $\mathbf{F}^m$ undergoes the global average pooling, obtaining the one-dimensional feature vector $\mathbf{f}^m \in \mathbb{R}^{c_m}$. Afterward, Z-score normalization \citep{al2006data} is applied to each feature vector $\mathbf{f}^{m}$, resulting in $\overline{\mathbf{f}}^{m} = \frac{\mathbf{f}^{m} - \mu_{\mathbf{f}^{m}}}{\sqrt{\operatorname{Var}(\mathbf{f}^{m}) + \delta}}$ for the $m$-th intermediate feature vector $\mathbf{f}^m$ of the input image $\mathbf{x}$, where $\operatorname{Var}(\mathbf{f}^{m})$ is the variance of $\mathbf{f}^{m}$ along the channel elements, and $\delta$ is a small constant value. Finally, we obtain the overall multi-layer feature vector for the input image $\mathbf{x}$ as: ${\mathbf z}_{0} = \mathcal{H}({\textbf{x}})  = [\overline{\mathbf{f}}^{1}, \ldots, \overline{\mathbf{f}}^{m}, \ldots, \overline{\mathbf{f}}^{M}] \in \mathbb{R}^c$ by concatenating all the intermediate feature vectors, where $c=\sum_{m=1}^{M} c_m$, and $\mathcal H(\mathbf x)$ denotes the whole feature extraction process.

\subsection{Diffusion-based Feature Distortion and Reconstruction}

\begin{wrapfigure}{r}{0.45\textwidth} % 'r' 代表右侧, 也可以用 'l' 代表左侧
  \centering
  \includegraphics[width=0.41\textwidth]{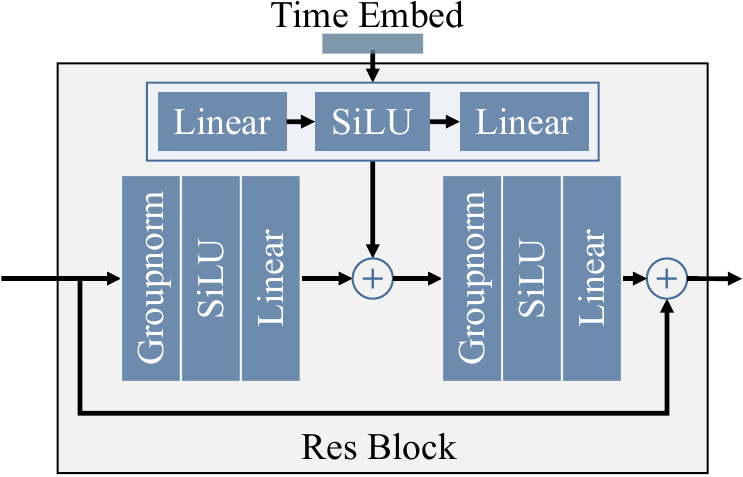} % 调整宽度和图片文件名
  \caption{Residual Block Structure in LFDN.}
   % \note{和图1联系不起来，图1中没有提及Time Embed}
  \label{fig:2}
\end{wrapfigure}
Fitting the semantic feature distribution of ID samples is crucial for identifying whether the input is an ID or OOD sample.
However, it is difficult to explicitly model the semantic feature space which has moderate complexity. 
Existing generative-based models \citep{zhou2022rethinking}, \citep{cai2023out} address the modeling of complex data/feature space by transferring the original data/features into an implicit bottleneck space and learning a generator capable of recovering ID samples from the bottleneck space.
Since the generator can not generalize well in recovering unseen OOD samples, it can be used as the OOD detector.
Inspired by this, we set up a diffusion-based feature distortion and reconstruction framework, considering the strength of diffusion models in data reconstruction.
Our framework is innovative in the introduction of diffusion models in modeling semantic features, while previous works \citep{graham2023denoising}, \citep{liu2023unsupervised}, \citep{gao2023diffguard} focus on applying diffusion models for straightforward pixel-level distortion and reconstruction.

\textbf{Semantic Feature Distortion.}

The semantic feature distortion process can be conceptualized as transforming the semantic features into distorted counterparts with different levels of noise. For each step \( t \) belonging to \( [1, \ldots, T] \), the generation of data point \(\mathbf{ z}_t \) follows the formula:
\begin{equation}
\mathbf{z}_t = \textrm{ennoise}(\mathbf{z}_0, t) = \sqrt{\overline{\alpha}_t} \times \mathbf{z}_0 + \sqrt{1-\overline{\alpha}_t} \times \boldsymbol{\epsilon}, \quad \boldsymbol{\epsilon} \sim \mathcal{N}(\mathbf 0^{c}, \mathbf I^{c\times c})
\end{equation}

where \( \boldsymbol{\epsilon} \in \mathbb R^c \) represents a Gaussian noise vector; \(\mathcal{N}(\cdot, \cdot )\) denotes the Gaussian distribution; $\mathbf 0^{c}$ and \( \mathbf I^{c\times c} \) denote the $c$-dimensional zero vector and  the $c\times c$ identity matrix, respectively. \( \overline{\alpha}_t \) is a predefined noise level that controls the amount of noise added at each step. 

\textbf{Semantic Feature Reconstruction.} For reconstructing the semantic features from their distorted counterparts, we build up a Latent Feature Diffusion Network (LFDN) constituted by 16 residual blocks (ResBlock), as shown in Fig.~\ref{fig:1}.

The structure of ResBlock is illustrated in Fig.~\ref{fig:2}. Its residual branch is formed with two groups of Groupnorm \citep{wu2018group}, SiLU, and linear layers, as well as a MLP used for absorbing in the time embedding.

Following the calculation process of the denoising diffusion implicit model~\citep{song2020denoising}, we employ LFDN to remove the noises injected into the semantic features with skipping step stride denoted as $s$. 
The detailed noise-removing process for $\mathbf z_t$  is described as follows. $s$ is set to a value randomly selected from $\{1,2,\cdots,t\}$. 

\begin{itemize}
\item[1)] We first input $\mathbf z_t$ and the time embedding of $t$ into LFDN, generating an initial reconstruction state denoted as $\tilde{\mathbf z}_t$. The calculation formulation can be summarized as: $\tilde{\mathbf z}_t = \textrm{LFDN}(\mathbf{z}_t, t)$, where $\textrm{LFDN}(\cdot)$ denotes the feed-forward process of LFDN. 

\item[2)] Afterwards, we estimate the noise correction vector for $\mathbf z_t$ denoted as $\tilde{\boldsymbol{\epsilon}}_t$ as follows,
\begin{equation}
\tilde{\boldsymbol{\epsilon}}_t = \frac{\mathbf{z}_t - \sqrt{\overline{\alpha}_t} \times \tilde{\mathbf z}_t}{\sqrt{1 - \overline{\alpha}_t}},
\end{equation}  
where \(\overline{\alpha}_t\) is the predefined noise level of the  \(t\)-th feature distortion step.

\item[3)] Then, we sample the input ($\tilde{\mathbf{z}}_{t^\prime}$) for implementing the $t^\prime$-th step's feature reconstruction where $t^\prime=\max(t-s,0)$  as:
\begin{equation}
\tilde{\mathbf{z}}_{t^\prime} = \sqrt{\overline{\alpha}_{t^\prime}} \left(\frac{\mathbf{z}_t - \sqrt{1 - \overline{\alpha}_t} \times \tilde{\boldsymbol{\epsilon}}_t)}{\sqrt{\overline{\alpha}_t}} + \sqrt{1 - \overline{\alpha}_{t^\prime} - \sigma_t^2} \times \tilde{\boldsymbol{\epsilon}}_t\right) + \sigma_t^2 \boldsymbol{\epsilon},
\label{eq:zt_minus_i}
\end{equation}
where \( \sigma_t^2 \) represents the variance of the additional noise at step \(t\).  Regarding $\tilde{\mathbf{z}}_{t^\prime}$ and time embedding of $t^\prime$ as inputs, LFDN predicts reconstruction results of the $t^\prime$-th step as $\tilde{\mathbf z}_{t^\prime} = \textrm{LFDN}(\tilde{\mathbf{z}}_{t^\prime}, t^\prime)$. 
\item[4)]Repeating steps 2 and 3 until $t^\prime$ = 0, yields the final reconstructed semantic features $\tilde{\mathbf z}_{0}$.
\end{itemize}

We summarize the above process as $\tilde{\mathbf z}_{0} = \textrm{denoise}(\mathbf z_t, t)$.
This framework ensures that \(\tilde{\mathbf z}_{0} \) is not solely derived from the LFDN output but is continuously refined by DDIM, integrating detailed corrections to achieve high accuracy in reconstructing the original data from its noisy observations.

\textbf{Objective Function}. 
For optimizing the network parameters of LFDN, the mean square error is used as the loss function for pulling close the outputs of LFDN with the original semantic features. The calculation formulation is as follows:
\begin{equation}
L = \frac{1}{N} \sum_{\mathbf x\in \mathbb D } \left\| \mathbf{z}_0 - \textrm{LFDN}(\mathbf{z}_{t}, t) \right\|_2^2 
\end{equation} 
During training, $t$ is randomly selected from $\{1,2,\cdots, T\}$. 
The detail is illustrated in Algorithm~\ref{alg:training}.

\subsection{OOD Detection Head}

Our approach can be integrated with three metrics to detect OOD data. Firstly, we utilize the Mean Squared Error (MSE) to measure the feature reconstruction error. Secondly, we use the Likelihood Regret metric (\(\text{LR} = \text{MSE}_{\text{initial}} - \text{MSE}_{\text{final}}\)) \citep{xiao2020likelihood}, which quantifies the change in MSE from the initial epoch to the final epoch. This metric reflects the model's evolving certainty during training. Generally, the reconstruction errors for ID samples decrease as the model becomes more familiar with these samples, whereas the errors for OOD samples remain relatively stable.
Lastly, we employ the Multi-layer Semantic Feature Similarity (\(\text{MFsim}\)), $i.e.$, the cosine similarity.
% The cosine similarity focuses on vector orientation similarity rather than magnitude, so it is more suitable for evaluating similarity in feature space. 
We assesses the cosine similarity between the original features  $\mathbf{z}_0 = [\overline{\mathbf{f}}^{1}, \ldots, \overline{\mathbf{f}}^{m}, \ldots, \overline{\mathbf{f}}^{M}]$ and the reconstructed features  $\tilde{\mathbf{z}}_0 = [\tilde{\mathbf{f}}^{1}, \ldots, \tilde{\mathbf{f}}^{m}, \ldots, \tilde{\mathbf{f}}^{M}]$ at various layers: \(\text{Sim}(\overline{\mathbf{f}}^{m}, \tilde{\mathbf{f}}^{m}) = \frac{\overline{\mathbf{f}}^{m} \cdot \tilde{\mathbf{f}}^{m}}{\|\overline{\mathbf{f}}^{m}\| \cdot\|\tilde{\mathbf{f}}^{m}\|}\). The OOD detection score \(\text{MFsim}\), is then computed as the negative average of these similarities: \(\text{MFsim}\) = $-\frac{1}{M} \sum_{m=1}^M \text{Sim}(\overline{\mathbf{f}}^{m}, \tilde{\mathbf{f}}^{m})$, where \(M\) is the number of feature maps. A higher \(\text{MFsim}\) score indicates a greater likelihood of the data being OOD.
Algorithm~\ref{alg:testing} details the MFsim calculation. The flows for MSE and LR calculations are provided in Appendix \ref{dddd}.

\vspace{-2mm}
\begin{figure}[htbp]
\centering
\begin{minipage}{.49\textwidth}
    \begin{algorithm}[H]
    \caption{Training Algorithm}
    \label{alg:training}
    \begin{algorithmic}[1] 
    \State \textbf{Input:} Train image \( \mathbf {x} \in \mathbb{R}^{3 \times h \times w} \)
    \State \( \mathbf{z}_0 = \mathcal{H}(\mathbf {x}) = [\overline{\mathbf{f}}^{1}, \ldots, \overline{\mathbf{f}}^{m}, \ldots, \overline{\mathbf{f}}^{M}] \in \mathbb{R}^c \)
    % \State \textbf{Output:} \( L_{\text{z}} \)
    \Repeat   

        \State Draw \( t \sim \text{Uniform}\{1, \ldots, T\} \)
       \vspace{0.2mm}
        \State Draw \( \epsilon \sim \mathcal{N}(0, I) \)
    \vspace{0.25mm}
        \State Compute \( \mathbf{z}_t \) and \( L \) 

        \State \( \mathbf{z}_t = \sqrt{\overline{\alpha}_t} \mathbf{z}_0 + \sqrt{1-\overline{\alpha}_t} \epsilon \)
  \vspace{0.25mm}
     % \State \( L_{\text{z}} = E_{\mathcal{H}(x), t} \left[ \left\|\mathbf{z}_{0} - LFDN(\mathbf{z}_{t}, t) \right\|_2^2 \right] \)
     \State \( L = \frac{1}{N} \sum_{\mathbf x\in \mathbb D } \left\|\mathbf{z}_{0} - \textrm{LFDN}(\mathbf{z}_{t}, t) \right\|_2^2 \)
      \vspace{0.25mm}
\State Update the parameters via the AdamW optimizer. %Take a numerical optimization step on \( \nabla_{\theta} L \) to 
\Until{convergence}
    \end{algorithmic}
    \end{algorithm}
\end{minipage}\hfill
\begin{minipage}{.48\textwidth}
    \begin{algorithm}[H]
    \caption{Testing Algorithm}
    \label{alg:testing}
    \begin{algorithmic}[1]  
    \State \textbf{Input:} An image \( \mathbf x \in \mathbb{R}^{3 \times h \times w}\)
    \State \textbf{Output:} OOD score
    %\For{\( i = 1 \) to \( n \)}
        \State \( \mathbf{z}_0 = \mathcal{H}(\mathbf x) = [\overline{\mathbf{f}}^{1}, \ldots, \overline{\mathbf{f}}^{m}, \ldots, \overline{\mathbf{f}}^{M}] \in \mathbb{R}^c \)
        \State \( \mathbf{z}_t \leftarrow \text{ennoise}(\mathbf{z}_0, t) \)
        \State \(  \tilde{\mathbf{z}}_0 \leftarrow \text{denoise}(\mathbf{z}_t, t) \)
        \State \(  [\tilde{\mathbf{f}}^{1}, \ldots, \tilde{\mathbf{f}}^{m}, \ldots, \tilde{\mathbf{f}}^{M}] \gets \tilde{\mathbf{z}}_0 \)
        \For{\( m = 1 \) to \( M \)}
            \State \( S_m \leftarrow \text{Sim}(\overline{\mathbf{f}}^{m}, \tilde{\mathbf{f}}^{m}) \)
        \EndFor
        \State \( \text{MFsim} \leftarrow -\left(\sum_{m=1}^M S_m \right)/M \)
    %\EndFor
    \State \Return \( \text{MFsim} \)
    \end{algorithmic}
    \end{algorithm}
\end{minipage}
\end{figure}

\section{Experiments} 
\label{Experiment}
\subsection{Datasets and Evaluation Metrics}
\textbf{Datasets}: We train the OOD detection model on three in-distribution (ID) datasets: CIFAR-10 \citep{krizhevsky2009learning}, CIFAR-100, and CelebA \citep{liu2015deep}. When testing models learned on a specific ID dataset, we select several datasets from SVHN \citep{netzer2011reading}, SUN \citep{xiao2010sun}, LSUN-c \citep{yu2015lsun}, LSUN-r, iSUN \citep{xu2015turkergaze}, iNaturalist \citep{van2018inaturalist}, Textures \citep{cimpoi2014describing}, Places365 \citep{zhou2017places}, MNIST \citep{deng2012mnist}, FMNIST, KMNIST \citep{clanuwat2018deep}, Omniglot \citep{lake2015human}, and NotMNIST as OOD data.

\textbf{Evaluation Metrics}: %For each input, our method outputs an OOD score, requiring a threshold for binary decision-making.  
We employed the area under the receiver operating characteristic (AUROC) and the false positive rate at 95\% true positive rate (FPR95) as evaluation metrics. Results in FPR95 metric are provided in Appendix~\ref{sec:fpr95_values}.

\subsection{Implementation Details}
\label{details}
We utilize EfficientNet-b4 \citep{tan2019efficientnet} or ResNet50 \citep{he2016deep}  pre-trained on ImageNet \citep{deng2009imagenet} as our encoder. 
The main text presents results using EfficientNet-b4, while results using ResNet50 are detailed in Appendix~\ref{sec:Resnet}. 
%Since the encoders are pre-trained on ImageNet, all training and testing images were resized to 224 × 224. 
For EfficientNet-b4, we select feature maps from the first to fifth stages (\(M=5\)) to construct the multi-layer semantic features, resulting in a feature dimension ($c$) of $720$. 
%where each stage represents a combination of blocks with similar dimensions.  
%These feature maps are concatenated to form a unified 720-dimensional single-vector feature, which is used as the input for the Latent Feature Diffusion Network (LFDN). 
The LFDN is consisting of 16 residual blocks. Inside each residual block, the number of groups in Groupnorm and the intermediate feature dimension of the residual branch are set to 1 and 1440, respectively. 
%For each ResBlock, the number of groups in GroupNorm, the number of residual blocks, and the dimensions of each linear layer are set to 1, 16, and 1440, respectively. 
%Finally, through a projection layer, it is transformed into a single-dimensional feature vector consistent with the original input dimension.
We employ the AdamW optimizer with a weight decay of \(10^{-4}\). Our method is trained on NVIDIA Geforce 4090 GPU for 150 epochs, with a batch size of 128 and a constant learning rate of \(10^{-4}\) throughout the training phase.

\subsection{Comparison with State-of-the-art Methods}

\textbf{Compared Generative-based Methods:} 
In Table~\ref{Table 1}, regarding CIFAR-10 as the ID dataset, we compare our method against pixel-level generative-based methods including GLOW~\citep{serra2019}, PixelCNN++~\citep{serra2019}, VAE~\citep{xiao2020likelihood}, and DDPM~\citep{graham2023denoising}. 
To validate the effectiveness of LFDN, we implement a variant of our method through replacing LFDN with AutoEncoder in which MFsim is used for estimating the OOD score.
In comparison with the best pixel-level method, VAE, our method achieves a 9.1\% improvement in average AUROC when using MFsim for OOD score estimation. 
Compared to DDPM, our method variants show a significantly improvement in average AUROC. For example, when integrated with MSE, our method achieves 20.4\% higher AUROC than DDPM.
%the three OOD score estimation strategies (MSE, LR, MFsim). 
This indirectly indicates that performing OOD detection at the pixel level is much worse than performing OOD detection at the feature level. 
Generating pixels may reconstruct more content unrelated to the image's semantics, which may interfere the identification of OOD samples.
Making the model focus on the reconstruction of compactly distributed semantic features benefits in separating ID and OOD samples. 
In terms of testing speed, our method is nearly 100 times faster than DDPM, significantly enhancing performance while reducing detection costs.
Moreover, the final version of our method built upon LFDN improves average AUROC by 18.5\% compared to the variant basd on AutoEncoder, as the diffusion model captures data distribution more effectively.  

In Table~\ref{Table 2}, we compare our method with VAE, DDPM and AutoEncoder, using CelebA as the ID dataset.
Our method integrated with MFsim achieves state-of-the-art performances, with an AUROC improvement of 19.89\% compared to DDPM, and the performance of the remaining two metrics also far exceeds the baseline, demonstrating the generalizability of our approach.

\textbf{Compared Classification-based and Distance-based Methods:} 
In Table~\ref{Table 3}, we compare our method with classification-based methods including MSP \citep{hendrycks2016baseline}, EBO \citep{liu2020energy}, DICE \citep{sun2022dice}, and ASH-S \citep{djurisic2022extremely}, as well as distance-based methods including `SimCLR+Mahalanobis Distance' \citep{xiao2021we} and `SimCLR+KNN' \citep{sun2022out}. All methods are evaluated using EfficientNet-b4 as the backbone. Compared to classification-based and distance-based methods, our approach consistently shows a clear advantage. Specifically, for CIFAR-100 as the in-distribution dataset, our method integrated with MFsim achieves an average AUROC of 13.84\% higher than the classification-based method DICE. Moreover, unlike classification-based methods, our approach does not require labeled data.

The inference speed of our method based on MSE or MFsim is faster than that of distance-based methods SimCLR+Maha and SimCLR+KNN, because the computation of covariance matrix or K nearest neighbors occupies part of time. Our method is also comparable to classifier-based methods including MSP, EBO, DICE and ASH-S.
This demonstrates the effectiveness of leveraging the strong ability of diffusion models to reconstruct original distributions from different noise levels for reconstructing low-dimensional features and performing OOD detection.

\begin{table}[htbp]
  \centering
  \caption{The AUROC values for OOD detection, where CIFAR-10 is used as the in-distribution dataset. The results are compared with generative-based methods. Higher AUROC values indicate better performance, with the best results highlighted in bold for clarity.}
  \resizebox{\textwidth}{!}{
    \begin{tabular}{cl|cccc|cccc}
    \toprule
    \multicolumn{2}{c|}{Dataset} & \multicolumn{4}{c|}{\textbf{Pixel-Generative-Base}} & \multicolumn{4}{c}{\textbf{Feature-Generative-Base}} \\
    \midrule
    ID    & \multicolumn{1}{c|}{OOD} & GLOW  & PixelCNN++ & VAE   & DDPM  & AutoEncoder & our(+MSE) & ours(+LR) & ours(+MFsim) \\
    \midrule
    \multirow{7}[2]{*}{CIFRA10} & \multicolumn{1}{c|}{SVHN} & 88.3  & 73.7  & 95.9  & 97.3  & 57.7  & 97.3±0.0 & 98.2±0.0 & \textbf{98.9±0.1} \\
          & \multicolumn{1}{c|}{LSUN} & 21.3  & 64.0  & 40.3  & 68.2  & 81.5  & 97.6±0.1 & 97.8±0.1 & \textbf{99.8±0.1} \\
          & \multicolumn{1}{c|}{MNIST} & 85.8  & 96.7  & \textbf{99.9} & 83.2  & 95.8  & 99.4±0.0 & 98.9±0.1 & \textbf{99.9±0.0} \\
          & \multicolumn{1}{c|}{FMNIST} & 71.2  & 90.7  & 99.1  & 84.3  & 79.6  & 99.0±0.0 & 98.8±0.0 & \textbf{99.9±0.0} \\
          & \multicolumn{1}{c|}{KMNIST} & 38.0  & 82.6  & \textbf{99.9} & 89.7  & 90.5  & 99.5±0.0 & 99.1±0.0 & \textbf{99.9±0.0} \\
          & \multicolumn{1}{c|}{Omniglot} & 95.5  & 98.9  & 99.6  & 35.9  & 81.5  & 99.1±0.1 & 97.1±0.1 & \textbf{99.9±0.0} \\
          & \multicolumn{1}{c|}{NotMNIST} & 53.9  & 82.6  & 99.4  & 88.7  & 81.6  & 99.8±0.1 & 99.5±0.0 & \textbf{99.9±0.0} \\
    \midrule
          & \multicolumn{1}{c|}{average} & 64.9  & 84.2  & 90.6  & 78.2 & 81.2 & 98.8±0.1 & 98.5±0.1 & \textbf{99.7±0.1} \\
    \midrule
    Time  & Num img/s (↑) & 38.6  & 19.3  & 0.7   & 11.4  & 1224.2 & 999.3 & 273.6 & 999.3 \\
    \bottomrule
    \end{tabular}%
    }
  \label{Table 1}%
\end{table}%

% Table generated by Excel2LaTeX from sheet 'Celeba'
\begin{table}[htbp]
  \centering
  \caption{The AUROC values for OOD detection, where CelebA is used as the in-distribution dataset. The results are compared with generative-based methods. Higher AUROC values indicate better performance, with the best results highlighted in bold for clarity.}
  \resizebox{\textwidth}{!}{
    \begin{tabular}{cc|cc|cccc}
    \toprule
    \multicolumn{2}{c|}{Dataset} & \multicolumn{2}{c|}{\textbf{Pixel-Generative-Based}} & \multicolumn{4}{c}{\textbf{Feature-Generative-Based}} \\
    \midrule
    ID    & OOD   & VAE   & DDPM  & AutoEncoder & ours(+MSE) & ours(+LR) & ours(+MFsim) \\
    \midrule
    \multirow{4}[2]{*}{CelebA} & SUN   & 95.89 & 83.41 & 32.90 & \textbf{99.98±0.01} & 97.15±0.02 & \textbf{99.98±0.01} \\
          & iNaturalist & 95.52 & 82.38 & 41.56 & \textbf{100+0.00} & 99.96±0.01 & 99.99±0.00 \\
          & Textures & 91.73 & 78.33 & 56.33 & 99.93±0.02 & 98.51±0.02 & \textbf{99.96±0.01} \\
          & Places365 & 97.58 & 76.25 & 35.90 & 99.96±0.01 & 97.47±0.03 & \textbf{99.98±0.00} \\
    \midrule
          & average & 95.18 & 80.09 & 41.67 & 99.97±0.01 & 98.27±0.02 & \textbf{99.98±0.01} \\
    \midrule
    Time  & Num img/s (↑) & 18.7  & 10.2  & 1357.6 & 1033.8 & 290.4 & 1033.8 \\
    \bottomrule
    \end{tabular}%
    }
  \label{Table 2}%
\end{table}%

% Table generated by Excel2LaTeX from sheet 'efficient'
\begin{table}[htbp]
  \centering
  \caption{The AUROC values for OOD detection, where CIFAR-10/100 is used as the in-distribution dataset. The results are compared with Classification-based and Distance-based methods using EfficientNet-b4 as the backbone. Higher AUROC values indicate better performance, with the best results highlighted in bold for clarity.}
  \resizebox{\textwidth}{!}{
    \begin{tabular}{c|c|l|r|cccccc|c}
    \toprule
    \multirow{2}[4]{*}{ID} & \multirow{2}[4]{*}{Based} & \multicolumn{1}{c|}{\multirow{2}[4]{*}{Method}} & \multicolumn{1}{c|}{\multirow{2}[4]{*}{Num img/s (↑)}} & \multicolumn{6}{c|}{OOD}          & \multirow{2}[4]{*}{\textbf{average }} \\
\cmidrule{5-10}        &     &     &     & SVHN & LSUN-c & LSUN-r & iSUN & Textures & Places365 &  \\
    \midrule
    \multirow{9}[6]{*}{CIFAR10} & \multirow{4}[2]{*}{Classifier-based } & MSP & 1060.5 & 94.53 & 96.37 & 91.80 & 92.23 & 95.93 & 97.59 & 94.74 \\
        &     & EBO & 1060.5 & 96.79 & 97.34 & 94.42 & 94.64 & 96.30 & 98.34 & 96.31 \\
        &     & DICE & 1066.3 & 98.53 & 99.03 & 94.49 & 95.25 & 97.68 & 99.63 & 97.44 \\
        &     & ASH-S & 1047.6 & 98.01 & 98.23 & 93.17 & 94.13 & 97.01 & 98.48 & 96.51 \\
\cmidrule{2-11}        & \multirow{2}[2]{*}{Distance-based} & SimCLR+Mahalanobis & 674.8 & 97.80 & 73.61 & 69.28 & 88.63 & 76.47 & 67.42 & 78.87 \\
        &     & SimCLR+KNN & 919.8 & 92.40 & 92.05 & 89.81 & 90.14 & 97.24 & 94.36 & 92.67 \\
\cmidrule{2-11}        & \multirow{3}[2]{*}{Generative-based} & ours(+MSE) & 960.6 & 97.31±0.02 & 97.59±0.01 & 93.93±0.01 & 92.78±0.01 & \textbf{100±0.00} & 99.96±0.00 & 96.93±0.01 \\
        &     & ours(+LR) & 360.2 & 98.22±0.02 & 97.84±0.02 & 95.37±0.01 & 94.31±0.02 & \textbf{100±0.00} & 99.91±0.01 & 97.61±0.02 \\
        &     & ours(+MFsim) & 960.6 & \textbf{98.89±0.01} & \textbf{99.83±0.02} & \textbf{98.83±0.01} & \textbf{98.52±0.02} & \textbf{100±0.00} & \textbf{100±0.00} & \textbf{99.34±0.01} \\
    \midrule
    \multirow{9}[6]{*}{CIFAR100} & \multirow{4}[2]{*}{Classifier-based } & MSP & 1060.5 & 77.56 & 84.03 & 72.09 & 71.52 & 90.02 & 89.00 & 80.70 \\
        &     & EBO & 1060.5 & 76.51 & 81.59 & 78.92 & 76.38 & 79.38 & 83.07 & 79.31 \\
        &     & DICE & 1066.3 & 86.93 & 88.54 & 71.97 & 71.29 & 92.83 & 90.78 & 83.72 \\
        &     & ASH-S & 1047.6 & 92.11 & 90.03 & 63.30 & 65.12 & 95.25 & 92.99 & 83.13 \\
\cmidrule{2-11}        & \multirow{2}[2]{*}{Distance-based} & SimCLR+Mahalanobis & 674.8 & 56.24 & 52.23 & 61.34 & 73.53 & 71.92 & 51.98 & 61.21 \\
        &     & SimCLR+KNN & 919.8 & 54.37 & 51.49 & 83.80 & 77.21 & 53.31 & 54.43 & 62.44 \\
\cmidrule{2-11}        & \multirow{3}[2]{*}{Generative-based} & ours(+MSE) & 960.6 & 83.93±0.01 & 86.86±0.01 & 75.38±0.01 & 71.99±0.02 & 99.99±0.00 & 99.97±0.01 & 86.35±0.01 \\
        &     & ours(+LR) & 360.2 & 88.84±0.01 & 87.60±0.02 & 80.96±0.01 & 77.71±0.02 & 99.98±0.01 & 99.92±0.02 & 89.17±0.01 \\
        &     & ours(+MFsim) & 960.6 & \textbf{93.90±0.01} & \textbf{99.14±0.01} & \textbf{95.74±0.01} & \textbf{94.40±0.01} & \textbf{100±0.00} & \textbf{100±0.00} & \textbf{97.20±0.01} \\
    \bottomrule
    \end{tabular}%
    }
  \label{Table 3}%
\end{table}%

\subsection{Ablation Study}

\begin{figure}[htbp]
    \centering
    \includegraphics[width=\linewidth]{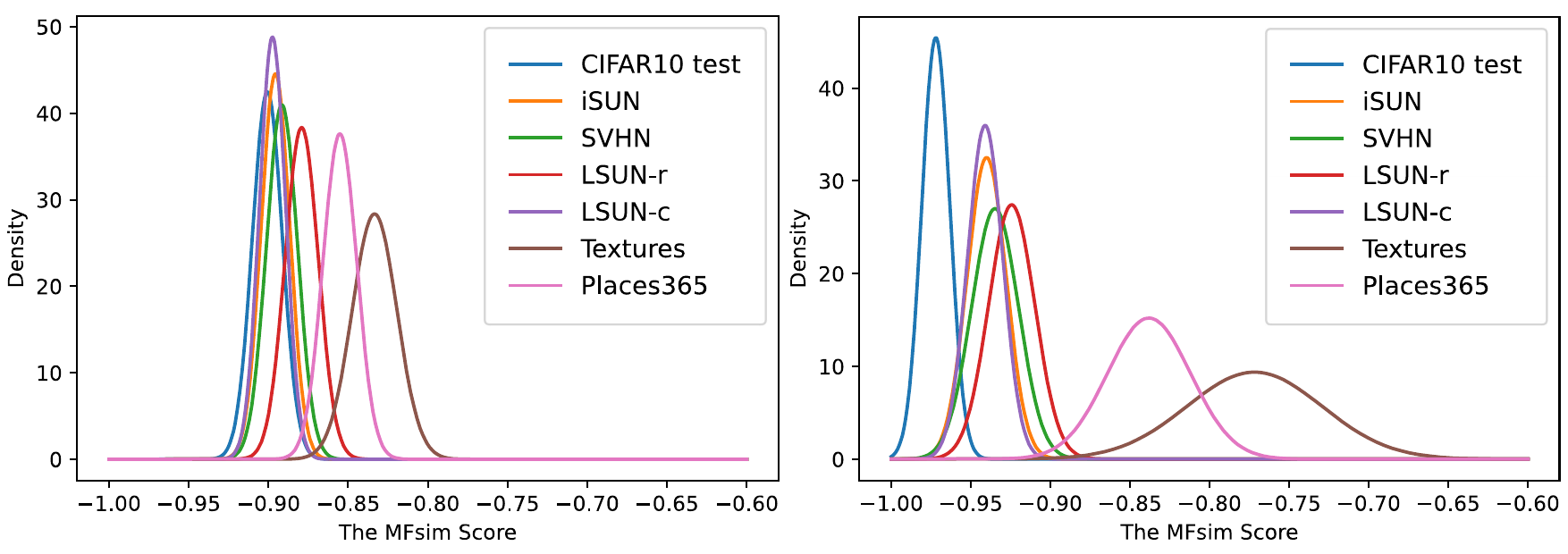} % 调整图片宽度以适合你的页面布局
    \caption{The MFsim score distributions of the first epoch (left) and the last epoch (right)}
    \label{fig:lft}
\end{figure}

\textbf{Illustration of the generation ability of the diffusion model on OOD detection}.
To demonstrate the evolution of the generative model's reconstruction capability for both ID and OOD samples before and after training, we compare the distributions of the MFsim scores at the first epoch and the final epoch in \textbf{Figure~\ref{fig:lft}}. CIFAR-10 serves as the ID dataset, while  the other six datasets listed in \textbf{Table~\ref{Table 3}} are employed as OOD data. Our observations reveal that the diffusion model's reconstruction ability enhances across most datasets, with a notably more pronounced improvement for the in-distribution samples. This indicates that ID samples are reconstructed more effectively, thereby validating the efficacy of our method.

\begin{figure}[htbp]
    \centering
    \includegraphics[width=0.9\linewidth]{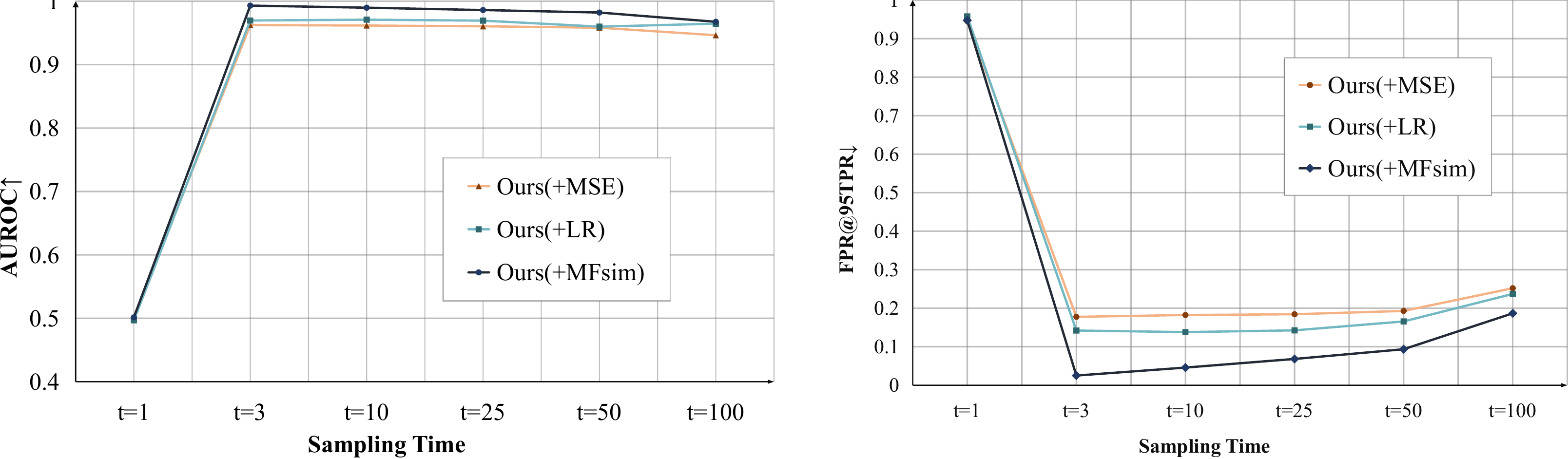} % 调整图片宽度以适合你的页面布局
    \caption{CIFAR-10 dataset is the ID data, the six datasets listed in Table 3 are used as OOD data. The average AUROC and FPR95  for the three metrics are evaluated at different sampling time steps.
}
    \label{fig:time_steps}
\end{figure}

\textbf{Performance variations across different sampling time steps}: \textbf{Figure\ ~\ref{fig:time_steps}} illustrates the variations in average AUROC and FPR95 values for different evaluation metrics at various sampling time steps, using CIFAR-10 as the ID data, with the final time step \( T = 100 \). It is observed that all metrics perform poorly at \( t = 1 \) primarily due to minimal noise added, making \( \mathbf{z}_t \) too similar to \( \mathbf{z}_0 \) and thus, limiting the denoising capability of LFDN; both ID and OOD data are well reconstructed. As \( t \) increases to about 3-10 steps, the appropriate amount of noise allows MSE, LR, and MFsim to reach optimal performances. However, as \( t \) continues to increase, the difference between \( \mathbf{z}_t \) and the original \( \mathbf{z}_0 \) enlarges, with \( \mathbf{z}_t \) gradually approaching random noise, thereby worsening the reconstruction differences between \( \tilde{\mathbf{z}}_0 \) and \( \mathbf{z}_0 \) for both ID and OOD samples.

% Table generated by Excel2LaTeX from sheet '线性层维度减半'
\begin{table}[htbp]
  \centering
  \caption{Changes in Average AUROC Across Six Datasets listed in Table 3 for CIFAR100 as ID.}
  \resizebox{\textwidth}{!}{
    \begin{tabular}{c|cc|cc|cc}
    \toprule
    \textbf{Metrics} & \multicolumn{2}{c|}{\textbf{MSE}} & \multicolumn{2}{c|}{\textbf{LR}} & \multicolumn{2}{c}{\textbf{MFsim}} \\
    \midrule
    Linear & Linear=720 & Linear=1440 & Linear=720 & Linear=1440 & Linear=720 & Linear=1440 \\
    \midrule
    \textbf{Average} & 83.35 & 86.35 & 84.05 & 89.17 & 96.43 & 97.20 \\
    \midrule
    Number of Blocks & Number=8 & Number=16 & Number=8 & Number=16 & Number=8 & Number=16 \\
    \midrule
    \textbf{Average} & 85.26 & 86.35 & 87.32 & 89.17 & 97.13 & 97.20 \\
    \bottomrule
    \end{tabular}%
    }
  \label{table 6}%
\end{table}%

\begin{figure}[htbp]
    \centering
    \includegraphics[width=\linewidth]{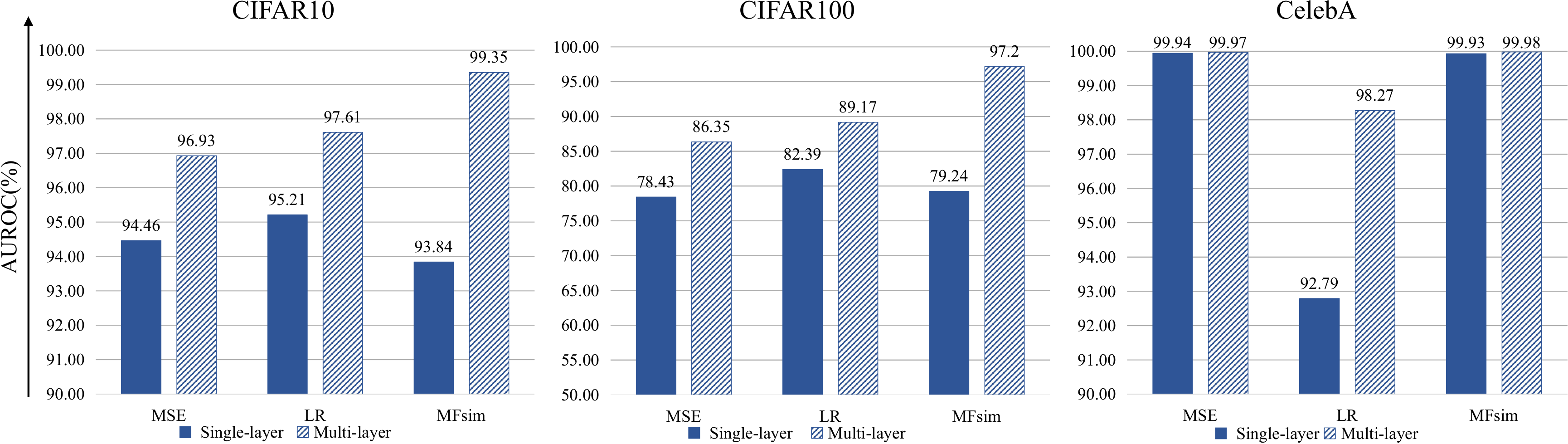}  % 确保文件名正确
    \caption{Variation of Average AUROC Values across Different Scales }
    \label{fig:multi_scale_auroc}
\end{figure}

\textbf{Comparison of MFsim across different feature scales}. \textbf{Figure\ ~\ref{fig:multi_scale_auroc}} displays performance comparisons of MFsim when reconstructing the last block (i.e., \( f_4, C=448 \)) versus multi-layer semantic features under an EfficientNet-b4 encoder. The results demonstrate that multi-layer semantic features generally outperform single-layer ones, indicating that multi-layer semantic features contain richer semantic information and are more representative of samples across different in-distribution datasets. Furthermore, considering the diverse semantic information represented by different layers, combing various layers of semantic features helps to boost the OOD performances of LFDN.
%we encourage future researchers to explore the impact of various feature combinations on the OOD detection performance of LFDN.

\textbf{Ablation study on LFDN network parameters}. 

We conducted ablation experiments on two groups of parameters within the LFDN network: the dimension of the linear layers and the number of ResBlocks. For each experiment, we reduced one of these parameters to half of its original size while keeping all other parameters unchanged. \textbf{Table~\ref{table 6}} presents the results of these experiments, showing how these modifications affect the performance. It is observed that the performance of our MFsim metric remains relatively stable, indicating that it continues to provide effective OOD detection capabilities even under conditions of reduced network size.
%\vspace{-5mm}
\section{Conclusion and Limitation}
\label{hahha}

In this paper, we propose a diffusion-based layer-wise semantic reconstruction framework for unsupervised OOD detection. We leverage the diffusion model's intrinsic data reconstruction ability to distinguish in-distribution and OOD samples in the latent feature space. Specially, the diffusion-based feature generation is built on top of the devised multi-layer semantic feature extraction strategy, which sets up a comprehensive and discriminative feature representation benefiting the generative OOD detection methods. Finally, we hope our proposed OOD detection method could make contributions to develop a safe real-world machine learning system. 
Additionally, it needs to point out that the performance of our method also relies on the quality of features extracted by the encoder. Therefore, selecting an encoder with strong feature extraction capabilities is crucial for achieving good performances.
 
\section{Acknowledgement}
This work was supported in part by the National Key R$\&$D Program of China under Grant No.2023YFA1008600, in part by NSFC under Grant NO.62376206, 62176198 and U22A2096,  in part by the Key R$\&$D Program of Shaanxi Province under Grant 2024GX-YBXM-135, in part by the Key Laboratory of Big Data Intelligent Computing under Grant BDIC-2023-A-004.
\section{Broader Impacts}
\label{eeesss}
\text{Positive Societal Impacts:} The proposed diffusion-based layer-wise semantic reconstruction method for unsupervised out-of-distribution (OOD) detection can significantly enhance the security and safety of machine learning systems. By effectively identifying OOD data, the system can prevent incorrect or potentially harmful decisions, making AI applications more reliable in critical areas such as healthcare, autonomous driving, and financial systems. This method increases the robustness of AI systems by ensuring they can handle unexpected inputs gracefully. This contributes to the overall stability and trustworthiness of AI deployments in various industries, thereby promoting wider acceptance and integration of AI technologies.
\text{Negative Societal Impacts}: As with any advanced detection method, there is a risk that the technology could be misused. For instance, surveillance applications, it could be employed to monitor individuals without their consent, leading to privacy violations and ethical concerns.
%\note{页数要求：The main text of a submitted paper is limited to nine content pages, including all figures and tables. Additional pages containing references don’t count as content pages.不能超页！}  
%\newpage
\bibliographystyle{unsrtnat} % or use neurips_2022 if it is defined
\bibliography{references}
%unsrt
% achemso
% plainnat
%plain

\newpage
%\vspace{+100mm}
\section*{Appendix} 

\appendix

\section{Supplementary Algorithm}
\label{dddd}

This section provides two key algorithms used for evaluating our approach: the MSE (Mean Squared Error) calculation and the Likelihood Regret (LR) calculation. 

The MSE calculation, as shown in Algorithm~\ref{alg:training3}, computes the mean squared error between the original and reconstructed latent features. It serves as a basic measure of reconstruction error for detecting OOD samples.

The LR calculation, detailed in Algorithm~\ref{alg:training4}, measures the reduction in reconstruction error by comparing the MSE values at the initial and final epochs of training. This metric reflects how well the model has adapted to the ID data over time, with a higher reduction indicating better adaptation.

\begin{minipage}{.98\textwidth}
    \begin{algorithm}[H]
    \caption{Testing Algorithm for MSE Calculation}
    \label{alg:training3}
    \begin{algorithmic}[1]
    \State \textbf{Input:} An image \( \mathbf x \)
    \State \textbf{Output:} MSE score
        \State \( \mathbf{z}_0 = \mathcal{H}(\mathbf x) \)
        \State \( \mathbf{z}_t \leftarrow \text{ennoise}(\mathbf{z}_0) \)
        \State \(  \tilde{\mathbf{z}}_0 = \text{denoise}(\mathbf{z}_t, t) \)
        \State \( \text{MSE} \leftarrow \frac{1}{N} \sum_{i=1}^N (\mathbf{z}_0[i] - \tilde{\mathbf{z}}_0[i])^2 \) \Comment{$i$ indexes the elements of $\mathbf{z}_0$ and $\tilde{\mathbf{z}}_0$}
    \State \Return \( \text{MSE} \)
    \end{algorithmic}
    \end{algorithm}
\end{minipage}

\begin{minipage}{.98\textwidth}
    \begin{algorithm}[H]
    \caption{Testing Algorithm for LR Calculation}
    \label{alg:training4}
    \begin{algorithmic}[1]
    \State \textbf{Input:} An image \( \mathbf x \), MSE at initial and final epochs
    \State \textbf{Output:} LR score
        \State \( \mathbf{z}_0^{\text{initial}} = \mathcal{H}(\mathbf x) \) at the beginning of training
        \State \( \mathbf{z}_t^{\text{initial}} \leftarrow \text{ennoise}(\mathbf{z}_0^{\text{initial}}) \)
        \State \( \tilde{\mathbf{z}}_0^{\text{initial}} = \text{denoise}(\mathbf{z}_t^{\text{initial}}, t) \)
        \State \( \text{MSE}_{\text{initial}} \leftarrow \frac{1}{N} \sum_{i=1}^N (\mathbf{z}_0^{\text{initial}}[i] - \tilde{\mathbf{z}}_0^{\text{initial}}[i])^2 \)
        
        \State \( \mathbf{z}_0^{\text{final}} = \mathcal{H}(\mathbf x) \) at the end of training
        \State \( \mathbf{z}_t^{\text{final}} \leftarrow \text{ennoise}(\mathbf{z}_0^{\text{final}}) \)
        \State \( \tilde{\mathbf{z}}_0^{\text{final}} = \text{denoise}(\mathbf{z}_t^{\text{final}}, t) \)
        \State \( \text{MSE}_{\text{final}} \leftarrow \frac{1}{N} \sum_{i=1}^N (\mathbf{z}_0^{\text{final}}[i] - \tilde{\mathbf{z}}_0^{\text{final}}[i])^2 \)
        
        \State \( \text{LR} \leftarrow \text{MSE}_{\text{initial}} - \text{MSE}_{\text{final}} \)
    \State \Return \( \text{LR} \)
    \end{algorithmic}
    \end{algorithm}
\end{minipage}

\section{More Experimental Details}

\subsection{Dataset Details and Testing Speeds}

\paragraph{\textbf{Table\ ~\ref{Table 1}} : CIFAR-10 Dataset}
The CIFAR-10 test set consisted of 10,000 images. The SVHN dataset contained 26,032 images, LSUN-r had 10,000 images, and Fashion-MNIST, MNIST, and KMNIST each comprised 10,000 images. Omniglot included 13,180 images, and notMNIST had 18,724 images, totaling 97,936 OOD samples. The testing of the MFsim metric took a total of 98 seconds, with an average speed of 999.3 images per second.

\paragraph{\textbf{Table\ ~\ref{Table 2}} : CelebA Dataset}
The CelebA test set comprised 60,780 images, SUN included 10,000 images, iNaturalist had 100,000 images, Textures consisted of 1,678 images, and Places365 had 1,002 images, making up a total of 112,680 OOD samples. Testing the MFsim metric took a total of 109 seconds, processing an average of 1033.8 images per second.

\subsection{Training Details}

Both CIFAR-10 and CelebA datasets were trained for 200 epochs using the VAE model. The GLOW model was trained for 150 epochs with a learning rate of $5 \times 10^{-4}$, and PixelCNN+ was trained for 150 epochs at the same learning rate. Under the DDPM model, both datasets were trained for 350 epochs, following the experimental setups and code provided in the original papers. We used LFDN without time-step embeddings as our autoencoder, used MFsim metrics, and kept all remaining training details consistent with our approach.

\section{Supplementary Experiments}

\subsection{Experimental Results for FPR95 Values Using EfficientNet-b4 as Backbone }
\label{sec:fpr95_values}
We conducted tests to evaluate the FPR95 (False Positive Rate at 95\% True Positive Rate) values using CIFAR10 and CIFAR100 datasets as in-distribution data while treating the remaining six datasets as out-of-distribution datasets. The specific FPR95 values are summarized in \textbf{Table\ ~\ref{Table7}}. 

\begin{table}[htbp]
  \centering
  \caption{The FPR95 values for OOD detection, where CIFAR-10/100 is used as the in-distribution dataset. The results are compared with Classification-based and Distance-based methods using EfficientNet-b4 as the backbone. Higher AUROC values indicate better performance, with the best results highlighted in bold for clarity.}
  \resizebox{\textwidth}{!}{
    \begin{tabular}{c|c|l|r|cccccc|c}
    \toprule
    \multirow{2}[4]{*}{ID} & \multirow{2}[4]{*}{Based} & \multicolumn{1}{c|}{\multirow{2}[4]{*}{Method}} & \multicolumn{1}{c|}{\multirow{2}[4]{*}{Num img/s (↑)}} & \multicolumn{6}{c|}{OOD}          & \multirow{2}[4]{*}{\textbf{average }} \\
\cmidrule{5-10}        &     &     &     & SVHN & LSUN-c & LSUN-r & iSUN & Textures & Places365 &  \\
    \midrule
    \multirow{9}[6]{*}{CIFAR10} & \multirow{4}[2]{*}{Classifier-based } & MSP & 1060.5 & 43.99 & 26.13 & 48.65 & 46.89 & 27.50 & 15.03 & 34.70 \\
        &     & EBO & 1060.5 & 16.51 & 11.52 & 28.38 & 27.03 & 16.08 & 4.99 & 17.42 \\
        &     & DICE & 1066.3 & 7.70 & 4.81 & 25.74 & 21.76 & 7.80 & 1.49 & 11.55 \\
        &     & ASH-S & 1047.6 & 6.89 & 4.15 & 31.29 & 26.29 & 5.21 & 1.32 & 12.53 \\
\cmidrule{2-11}        & \multirow{2}[2]{*}{Distance-based} & SimCLR+Mahalanobis & 674.8 & 9.24 & 67.73 & 75.43 & 64.32 & 56.22 & 72.15 & 57.52 \\
        &     & SimCLR+KNN & 919.8 & 49.15 & 54.89 & 76.97 & 73.48 & 15.27 & 39.39 & 51.53 \\
\cmidrule{2-11}        & \multirow{3}[2]{*}{Generative-based} & ours(+MSE) & 960.6 & 21.15±0.03 & 19.52±0.01 & 39.67±0.02 & 43.76±0.02 & \textbf{0±0.00} & 0.42±0.03 & 20.75±0.02 \\
        &     & ours(+LR) & 360.2 & 14.26±0.02 & 18.67±0.03 & 31.62±0.02 & 37.76±0.02 & 0.06±0.01 & 0.83±0.02 & 17.20±0.02 \\
        &     & ours(+MFsim) & 960.6 & \textbf{4.34±0.02} & \textbf{0.04±0.01} & \textbf{4.42±0.02} & \textbf{6.26±0.02} & \textbf{0±0.00} & \textbf{0±0.00} & \textbf{2.51±0.01} \\
    \midrule
    \multirow{9}[6]{*}{CIFAR100} & \multirow{4}[2]{*}{Classifier-based } & MSP & 1060.5 & 80.10 & 68.80 & 80.35 & 80.36 & 47.11 & 57.41 & 69.02 \\
        &     & EBO & 1060.5 & 88.74 & 78.64 & 72.35 & 76.57 & 95.83 & 94.04 & 84.36 \\
        &     & DICE & 1066.3 & 63.77 & 58.96 & 77.89 & 78.67 & 34.26 & 48.77 & 60.39 \\
        &     & ASH-S & 1047.6 & \textbf{34.28} & 44.39 & 89.45 & 86.13 & 21.44 & 34.59 & 51.71 \\
\cmidrule{2-11}        & \multirow{2}[2]{*}{Distance-based} & SimCLR+Mahalanobis & 674.8 & 94.95 & 96.35 & 85.05 & 86.29 & 80.37 & 95.50 & 89.75 \\
        &     & SimCLR+KNN & 919.8 & 95.32 & 97.11 & 78.45 & 84.38 & 95.28 & 94.82 & 90.89 \\
\cmidrule{2-11}        & \multirow{3}[2]{*}{Generative-based} & ours(+MSE) & 960.6 & 89.05±0.02 & 69.14±0.03 & 69.63±0.02 & 89.79±0.01 & 0.12±0.02 & \textbf{0±0.00} & 52.95±0.02 \\
        &     & ours(+LR) & 360.2 & 62.06±0.03 & 72.19±0.04 & 86.67±0.02 & 86.17±0.02 & 0.42±0.02 & 2.81±0.02 & 51.72±0.03 \\
        &     & ours(+MFsim) & 960.6 & 37.48±0.02 & \textbf{1.90±0.01} & \textbf{23.05±0.02} & \textbf{26.00±0.02} & \textbf{0±0.00} & \textbf{0±0.00} & \textbf{14.78±0.02} \\
    \bottomrule
    \end{tabular}%
    }
  \label{Table7}%
\end{table}%

As shown in \textbf{Table~\ref{Table7}}, our method demonstrates a significant advantage in terms of FPR95 values compared to other classification-based and distance-based approaches. Specifically, when using CIFAR100 as in-distribution data, our method achieves an average reduction of 36.93\% in FPR95 values compared to the state-of-the-art classification-based approach, ASH-S.

\subsection{ Experimental Results with ResNet50 as Encoder}
\label{sec:Resnet}
Besides using EfficientNet-b4 as the encoder, we also employed the commonly used ResNet50 to extract multi-layer semantic features. For ResNet50, feature maps from stages 1 to 3 were selected, with channel counts of 256, 512, and 1024, respectively. These feature maps were concatenated to form a 1792-dimensional vector, which was then used as input for the LFDN. The results for three OOD detection metrics are presented in \textbf{Table~\ref{Table8}} and \textbf{Table~\ref{table9}}. Both tables compare our method with classification-based and distance-based methods.

% Table generated by Excel2LaTeX from sheet 'resnet50'
\begin{table}[htbp]
  \centering
  \caption{The AUROC values for OOD detection, where CIFAR-10/100 is used as the in-distribution dataset. The results are compared with Classification-based and Distance-based methods using ResNet50 as the backbone. Higher AUROC values indicate better performance, with the best results highlighted in bold for clarity.}
  \resizebox{\textwidth}{!}{
    \begin{tabular}{c|c|l|r|cccccc|c}
    \toprule
    \multirow{2}[4]{*}{ID} & \multirow{2}[4]{*}{Based} & \multicolumn{1}{c|}{\multirow{2}[4]{*}{Method}} & \multicolumn{1}{c|}{\multirow{2}[4]{*}{Num img/s (↑)}} & \multicolumn{6}{c|}{OOD}          & \multirow{2}[4]{*}{\textbf{average}} \\
\cmidrule{5-10}        &     &     &     & SVHN & LSUN-c & LSUN-r & iSUN & Textures & Places365 &  \\
    \midrule
    \multirow{9}[6]{*}{CIFAR10} & \multirow{4}[2]{*}{Classifier-based } & MSP & 1321.5 & 80.64 & 95.05 & 91.65 & 89.95 & 87.78 & 90.43 & 89.25 \\
        &     & EBO & 1321.5 & 81.90 & 98.21 & 94.43 & 93.09 & 86.84 & 92.45 & 91.15 \\
        &     & DICE & 1369.3 & 91.92 & 99.18 & 91.13 & 90.51 & 87.39 & 92.66 & 92.13 \\
        &     & ASH-S & 1307.4 & 84.16 & 98.76 & 95.00 & 94.40 & 87.88 & 91.63 & 91.97 \\
\cmidrule{2-11}        & \multirow{2}[2]{*}{Distance-based} & SimCLR+Mahalanobis & 857.2 & 90.17 & 73.74 & 86.14 & 83.25 & 81.48 & 91.23 & 84.34 \\
        &     & SimCLR+KNN & 1179.8 & 92.36 & 91.13 & 87.78 & 91.82 & 88.62 & 79.41 & 88.52 \\
\cmidrule{2-11}        & \multirow{3}[2]{*}{Generative-based} & ours(+MSE) & 654.9 & 81.40 & 94.02 & 81.11 & 81.36 & 100.00 & 100.00 & 89.65 \\
        &     & ours(+LR) & 296.7 & 90.95 & 92.53 & 85.22 & 85.91 & 100.00 & 100.00 & 92.44 \\
        &     & ours(+MFsim) & 654.9 & \textbf{95.98±0.02} & \textbf{99.86±0.02} & \textbf{97.06±0.02} & \textbf{96.89±0.01} & \textbf{100±0.00} & \textbf{100±0.00} & \textbf{98.30±0.01} \\
    \midrule
    \multirow{9}[6]{*}{CIFAR100} & \multirow{4}[2]{*}{Classifier-based } & MSP & 1321.5 & 78.38 & 84.18 & 78.98 & 78.09 & 76.54 & 72.00 & 78.03 \\
        &     & EBO & 1321.5 & 83.13 & 89.35 & 83.83 & 83.35 & 78.85 & 71.42 & 81.66 \\
        &     & DICE & 1369.3 & 87.93 & 93.32 & 82.41 & 82.20 & 79.29 & 69.65 & 82.47 \\
        &     & ASH-S & 1307.4 & 91.66 & 93.24 & 69.93 & 72.78 & 87.75 & 71.00 & 81.06 \\
\cmidrule{2-11}        & \multirow{2}[2]{*}{Distance-based} & SimCLR+Mahalanobis & 857.2 & \textbf{91.92} & 57.14 & 87.47 & 88.00 & 94.96 & 71.86 & 81.89 \\
        &     & SimCLR+KNN & 1179.8 & 87.78 & 84.30 & 82.51 & 77.69 & 83.35 & 73.74 & 81.56 \\
\cmidrule{2-11}        & \multirow{3}[2]{*}{Generative-based} & ours(+MSE) & 654.9 & 86.55 & 99.11 & 93.02 & 91.87 & 100.00 & 100.00 & 95.09  \\
        &     & ours(+LR) & 296.7 & 89.17 & 99.16 & \textbf{93.65} & 92.33 & 100.00 & 100.00 & 95.72  \\
        &     & ours(+MFsim) & 654.9 & 89.68±0.02 & \textbf{99.18±0.01} & 93.64±0.02 & \textbf{92.94±0.02} & \textbf{100±0.00} & \textbf{100±0.00} & \textbf{95.91±0.01} \\
    \bottomrule
    \end{tabular}%
    }
  \label{Table8}%
\end{table}%

% Table generated by Excel2LaTeX from sheet 'resnet50'
% Table generated by Excel2LaTeX from sheet 'Sheet3'
\begin{table}[htbp]
  \centering
  \caption{The FPR95 values for OOD detection, where CIFAR-10/100 is used as the in-distribution dataset. The results are compared with Classification-based and Distance-based methods using ResNet50 as the backbone. Higher AUROC values indicate better performance, with the best results highlighted in bold for clarity.}
  \resizebox{\textwidth}{!}{
    \begin{tabular}{c|c|l|r|cccccc|c}
    \toprule
    \multirow{2}[4]{*}{ID} & \multirow{2}[4]{*}{Based} & \multicolumn{1}{c|}{\multirow{2}[4]{*}{Method}} & \multicolumn{1}{c|}{\multirow{2}[4]{*}{Num img/s (↑)}} & \multicolumn{6}{c|}{OOD}          & \multirow{2}[4]{*}{\textbf{average}} \\
\cmidrule{5-10}        &     &     &     & SVHN & LSUN-c & LSUN-r & iSUN & Textures & Places365 &  \\
    \midrule
    \multirow{9}[5]{*}{CIFAR10} & \multirow{4}[2]{*}{Classifier-based } & MSP & 1321.5 & 60.56 & 16.09 & 28.21 & 35.63 & 45.4 & 32.93 & 36.47 \\
        &     & EBO & 1321.5 & 59.15 & 8.07 & 23.79 & 30.72 & 60.2 & 31.66 & 35.60 \\
        &     & DICE & 1369.3 & 27.07 & 4.11 & 37.83 & 41.12 & 57.48 & 32.32 & 33.32 \\
        &     & ASH-S & 1307.4 & 53.93 & 5.57 & 20.29 & 22.31 & 64.36 & 39.71 & 34.36 \\
\cmidrule{2-11}        & \multirow{2}[2]{*}{Distance-based} & SimCLR+Mahalanobis & 857.2 & 27.65 & 33.35 & 48.17 & 51.22 & 38.12 & 60.43 & 43.16 \\
        &     & SimCLR+KNN & 1179.8 & 24.53 & 25.29 & 37.81 & 27.55 & 34.36 & 62.19 & 35.29 \\
\cmidrule{2-11}        & \multirow{3}[1]{*}{Generative-based} & ours(+MSE) & 654.9 & 94.79 & 45.99 & 95.86 & 91.68 & 0.00 & 0.00 & 54.72 \\
        &     & ours(+LR) & 296.7 & 34.59 & 25.77 & 56.54 & 52.05 & 0.00 & 0.00 & 28.16 \\
        &     & ours(+MFsim) & 654.9 & \textbf{21.10±0.03} & \textbf{0.02±0.01} & \textbf{15.75±0.03} & \textbf{16.48±0.04} & \textbf{0.00±0.00} & \textbf{0.00±0.00} & \textbf{8.89±0.02} \\
         \midrule
    \multirow{9}[5]{*}{CIFAR100} & \multirow{4}[1]{*}{Classifier-based } & MSP & 1321.5 & 53.38  & 43.52  & 56.23  & 57.69  & 63.63  & 77.53  & 58.66 \\
        &     & EBO & 1321.5 & 47.04 & 34.15 & 51.14 & 52.36 & 63.05 & 80.95 & 54.78 \\
        &     & DICE & 1369.3 & 38.7 & 28.77 & 56.21 & 56.74 & 65.21 & 82.63 & 54.71 \\
        &     & ASH-S & 1307.4 & \textbf{29.83} & 28.75 & 89.48 & 85.22 & 51.8 & 81.48 & 61.09 \\
\cmidrule{2-11}        & \multirow{2}[2]{*}{Distance-based} & SimCLR+Mahalanobis & 857.2 & 32.19 & 80.43 & 39.93 & 40.39 & 28.21 & 81.44 & 50.43 \\
        &     & SimCLR+KNN & 1179.8 & 39.23 & 48.99 & 60.21 & 74.99 & 57.15 & 80.74 & 60.22 \\
\cmidrule{2-11}        & \multirow{3}[2]{*}{Generative-based} & ours(+MSE) & 654.9 & 78.54 & \textbf{0.49} & 41.81 & 43.12 & 0.00 & 0.00 & 27.33 \\
        &     & ours(+LR) & 296.7 & 65.93 & 0.57 & \textbf{36.71} & \textbf{38.78} & 0.00 & 0.00 & \textbf{23.67} \\
        &     & ours(+MFsim) & 654.9 & 64.72±0.03 & 1.08±0.02 & 39.66±0.04 & 39.79±0.03 & \textbf{0.00±0.00} & \textbf{0.00±0.00} & 24.21±0.02 \\
    \bottomrule
    \end{tabular}%
    }
  \label{table9}%
\end{table}%

As shown in \textbf{Table~\ref{Table8}} and \textbf{Table~\ref{table9}}, when using ResNet50 as the backbone, our method still achieves the best performance. Specifically, with CIFAR-10 as the in-distribution dataset, the average AUROC and MFsim values are 98.30\% and 8.89\%, respectively, outperforming the classification-based SOTA method DICE by 6.17\% in AUROC and reducing the FPR95 by 24.43\%.

\begin{figure}[htbp]
    \centering
    \begin{minipage}{0.48\linewidth}
        \centering
        \includegraphics[width=\linewidth]{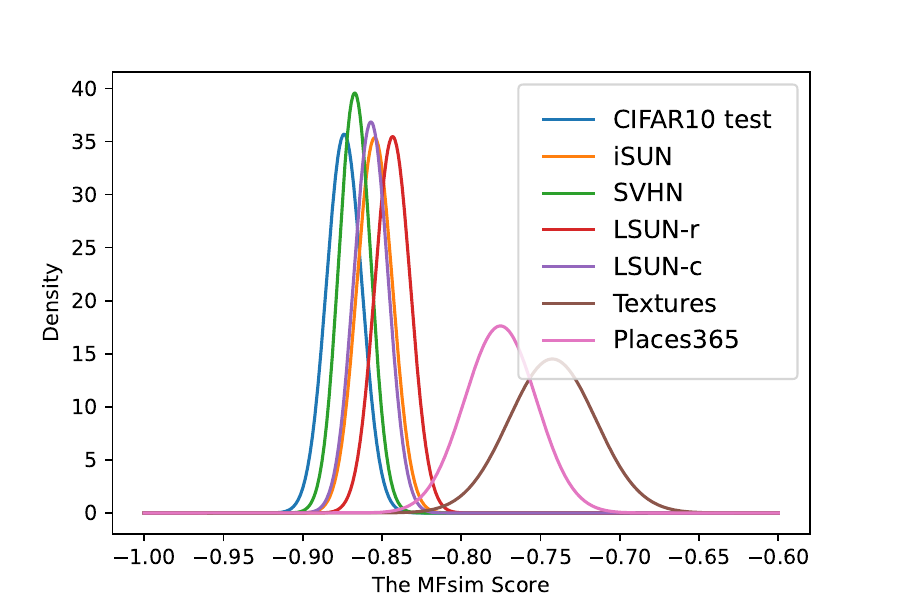}
        \caption{The MFsim score distributions of the First Epoch with ResNet50 as Encoder}
        \label{first_epoch1111}
    \end{minipage}
    \hfill
    \begin{minipage}{0.48\linewidth}
        \centering
        \includegraphics[width=\linewidth]{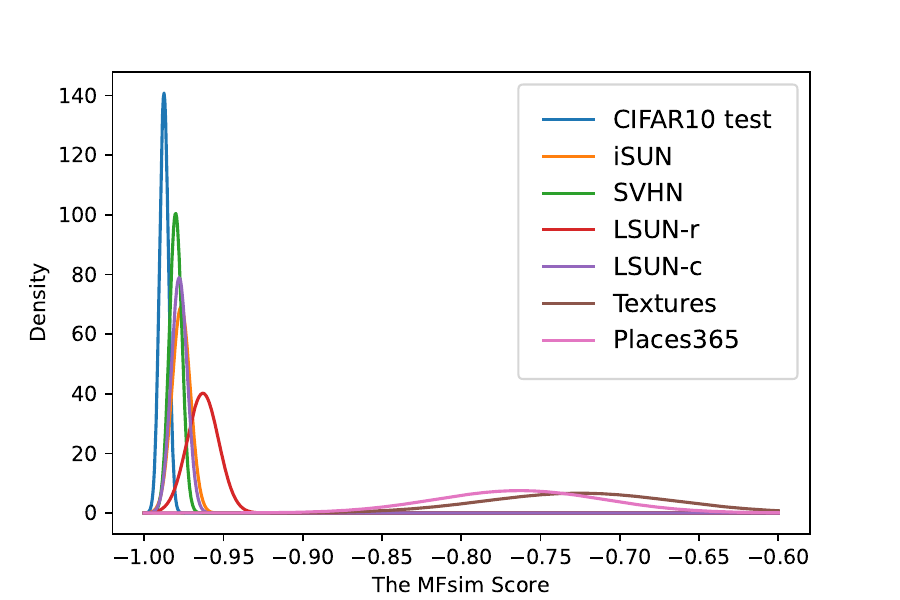}
        \caption{The MFsim score distributions of the Last Epoch with ResNet50 as Encoder}
        \label{last_epoch}
    \end{minipage}
    % \caption{The MFsim score distributions of the First and Last Epochs with ResNet50 as Encoder}
    \label{combined_distributions1111}
\end{figure}

Figures \ref{first_epoch1111} and \ref{last_epoch} illustrate the differences in the MFsim score distributions for various datasets, with ResNet50 as the encoder and CIFAR10 as the in-distribution dataset, across the first and last epochs.

% \begin{figure}[htbp]
%     \centering
%     \includegraphics[width=\linewidth]{combined_distributions4resnet50fistt=10.pdf} % 调整图片宽度以适合你的页面布局
%     \caption{The MFsim score distributions of the First Epoch with ResNet50 as Encoder}
%     \label{first epoch}
% \end{figure}

% \begin{figure}[htbp]
%     \centering
%     \includegraphics[width=\linewidth]{combined_distributions4resnet50lastt=10.pdf} % 调整图片宽度以适合你的页面布局
%     \caption{The MFsim score distributions of the last Epoch with ResNet50 as Encoder}
%     \label{last epoch}
% \end{figure}

\subsection{CIFAR-10 as ID and CIFAR-100 as OOD}
As shown in \textbf{Table~\ref{table10}}, when CIFAR-10 is used as the ID dataset and CIFAR-100 as the OOD dataset, our method consistently achieves the best performance across different evaluation metrics, including AUROC and FPR95. Compared to classification-based methods, the improvement is not significantly large, but our approach still shows a consistent edge, particularly in feature-generative-based models, demonstrating the robustness of our method.

\begin{table}[htbp]
  \centering
  \caption{The FPR95 and AUROC Values for CIFAR-10 as ID Samples and CIFAR100 as OOD Samples.}
  \resizebox{\textwidth}{!}{
    \begin{tabular}{c|c|c|l|c|c}
    \toprule
    ID  & OOD & \multicolumn{1}{l|}{Based} & Method & FPR95↓ & AUROC↑ \\
    \midrule
    \multirow{9}[6]{*}{CIFAR10} & \multirow{9}[6]{*}{CIFAR100} & \multirow{3}[2]{*}{Classification-Based} & MSP & 52.04 & 86.14 \\
        &     &     & EBO & 51.32 & 86.19 \\
        &     &     & ASH-S & 51.29 & 87.13 \\
    \cmidrule{3-6}        
        &     & \multirow{3}[2]{*}{Pixel-Generative-Based} & GLOW & -   & 73.60 \\
        &     &     & VAE & 90.41 & 55.95 \\
        &     &     & DDPM & 67.38 & 82.43 \\
    \cmidrule{3-6}        
        &     & \multirow{3}[2]{*}{Feature-Generative-Based} & ours(+MSE) & \textbf{48.87} & \textbf{87.54} \\
        &     &     & ours(+LR) & 49.48 & 87.24 \\
        &     &     & ours(+MFsim) & 53.70 & 85.60 \\
    \bottomrule
    \end{tabular}%
  }
  \label{table10}%
\end{table}%

% \end{table}%

\subsection{Comparisons with recent generative methods}
The comparison between our method and DDPM\citep{graham2023denoising} can be referred to \textbf{Table~\ref{Table 1}} and \textbf{Table~\ref{Table 2}}. Our method outperforms DDPM consistently on benchmarks using CIFAR10 or CelebA as ID data. 

The comparison between our method and Diffuard\citep{gao2023diffguard} is provided in \textbf{Table~\ref{table12}}. Results of Diffuard are taken from its original paper. Here, CIFAR10 is regarded as ID data, while CIFAR100 or TinyImagenet is regarded as OOD data. Our method based on MFsim achieves overall better performance than ‘Diffuard+Deep Ens’, with 1.55 higher AUROC and 21.77 lower FPR95.

The comparison between our method and LMD\citep{liu2023unsupervised} is shown in \textbf{Table~\ref{table13}}. The evaluation metric is AUROC. The average AUROC of our method based on MFsim is 6.94 higher than that of LMD.
% Table generated by Excel2LaTeX from sheet '与最新生成基比较'
\begin{table}[htbp]
  \centering
  \caption{The AUROC and FPR95 values compared to DiffGuard \citep{gao2023diffguard} using CIFAR-10 as the ID dataset and CIFAR-100/TinyImageNet as the OOD datasets.}
   \resizebox{\textwidth}{!}{
    \begin{tabular}{l|rr|rr|rr|}
    \toprule
    \multicolumn{1}{c|}{\multirow{2}[4]{*}{Method}} & \multicolumn{1}{c}{CIFAR-100 } &     & \multicolumn{2}{c|}{TINYIMAGENET} & \multicolumn{2}{c}{average} \\
\cmidrule{2-7}        & \multicolumn{1}{l}{AUROC↑} & \multicolumn{1}{l|}{FPR95↓} & \multicolumn{1}{l}{AUROC↑} & \multicolumn{1}{l|}{FPR95↓} & \multicolumn{1}{l}{AUROC↑} & \multicolumn{1}{l}{FPR95↓} \\
    \midrule
    Diffuard & 89.88 & 52.67 & 91.88 & 45.48 & 90.88 & 49.08 \\
    Diffuard+EBO & 89.93 & 50.77 & 91.95 & 43.58 & 90.94 & 47.18 \\
    Diffuaed+Deep Ens & \textbf{90.40} & 52.51 & 91.98 & 45.04 & 91.19 & 48.78 \\
    \midrule
    ours(+MSE) & 87.54 & \textbf{48.87} & 97.68 & 13.42 & 92.61 & 31.15 \\
    ours(+LR) & 87.24 & 49.48 & 97.11 & 15.04 & 92.18 & 32.26 \\
    ours(+MFsim) & 85.60 & 53.70 & \textbf{99.88} & \textbf{0.39} & \textbf{92.74} & \textbf{27.01} \\
    \bottomrule
    \end{tabular}%
    }
  \label{table12}%
\end{table}%

% Table generated by Excel2LaTeX from sheet '与最新生成基比较'
% \begin{table}[htbp]
%   \centering
%   \caption{The AUROC values compared to LMD \citep{liu2023unsupervised} using CIFAR-10/CIFAR-100 as the ID dataset and CIFAR-100/CIFAR10/SVHN as the OOD datasets.}
%   \resizebox{\textwidth}{!}{
%     \begin{tabular}{cc|r|rrr}
%     \toprule
%     \multicolumn{1}{c|}{ID} & \multicolumn{1}{l|}{OOD} & \multicolumn{1}{l|}{LMD} & \multicolumn{1}{l}{ours(+MSE)} & \multicolumn{1}{l}{ours(+LR)} & \multicolumn{1}{l}{ours(+MFsim)} \\
%     \midrule
%     \multicolumn{1}{c|}{\multirow{2}[1]{*}{CIFAR10}} & \multicolumn{1}{l|}{CIFAR100} & 60.70 & 87.54 & 87.24 & 85.60 \\
%     \multicolumn{1}{c|}{} & \multicolumn{1}{l|}{SVHN} & 99.20 & 97.31 & 98.22 & 98.89 \\
%     \multicolumn{1}{c|}{\multirow{2}[1]{*}{CIFAR100}} & \multicolumn{1}{l|}{CIFAR10} & 56.80 & 70.52 & 72.86 & 64.58 \\
%     \multicolumn{1}{c|}{} & \multicolumn{1}{l|}{SVHN} & 98.50 & 83.93 & 88.84 & 93.90 \\
%     \midrule
%     \multicolumn{2}{c|}{AVERAGE} & 78.80 & 84.83 & \textbf{86.79} & 85.74 \\
%     \bottomrule
%     \end{tabular}%
%     }
%   \label{table13}%
% \end{table}%
% Table generated by Excel2LaTeX from sheet '与最新生成基比较'
\begin{table}[htbp]
  \centering
  \caption{The AUROC values compared to LMD \citep{liu2023unsupervised} using CIFAR-10/CIFAR-100 as the ID dataset and CIFAR-100/CIFAR10/SVHN as the OOD datasets.}
  \resizebox{\textwidth}{!}{
    \begin{tabular}{cc|c|c|c|c}
    \toprule
    \multicolumn{1}{c|}{ID} & OOD & LMD & ours(+MSE) & ours(+LR) & ours(+MFsim) \\
    \midrule
    \multicolumn{1}{c|}{\multirow{2}[2]{*}{CIFAR10}} & CIFAR100 & 60.70 & 87.54 & 87.24 & 85.6 \\
    \multicolumn{1}{c|}{} & SVHN & 99.20 & 97.31 & 98.22 & 98.89 \\

    \multicolumn{1}{c|}{\multirow{2}[2]{*}{CIFAR100}} & CIFAR10 & 56.80 & 70.52 & 72.86 & 64.58 \\
    \multicolumn{1}{c|}{} & SVHN & 98.50 & 83.93 & 88.84 & 93.9 \\
    \midrule
    \multicolumn{2}{c|}{AVERAGE} & 78.80 & 84.83 & \textbf{86.79} & 85.74 \\
    \bottomrule
    \end{tabular}%
    }
  \label{table13}%
\end{table}%

\subsection{ Comparison against other methods using the multi-scale feature encodings as the input.}
In \textbf{Table~\ref{tablle17}}, we have made comparison of our method against AE and VAE using the multi-layer feature encodings as inputs.
For AE (AutoEncoder), we use the LFDN network without the timestep embedding, i.e., a 16-layer linear network. For VAE, we use a 5-layer linear network as the encoder and an 8-layer linear network as the decoder.
Compared to AE and VAE, the diffusion model has significant advantages when modeling complex multidimensional distributions. 
\subsection{Comparisons with pixel-level denoising approaches.}
\begin{figure}[htbp]
    \centering
    \begin{minipage}{0.48\linewidth}
        \centering
        \includegraphics[width=\linewidth, height=4cm]{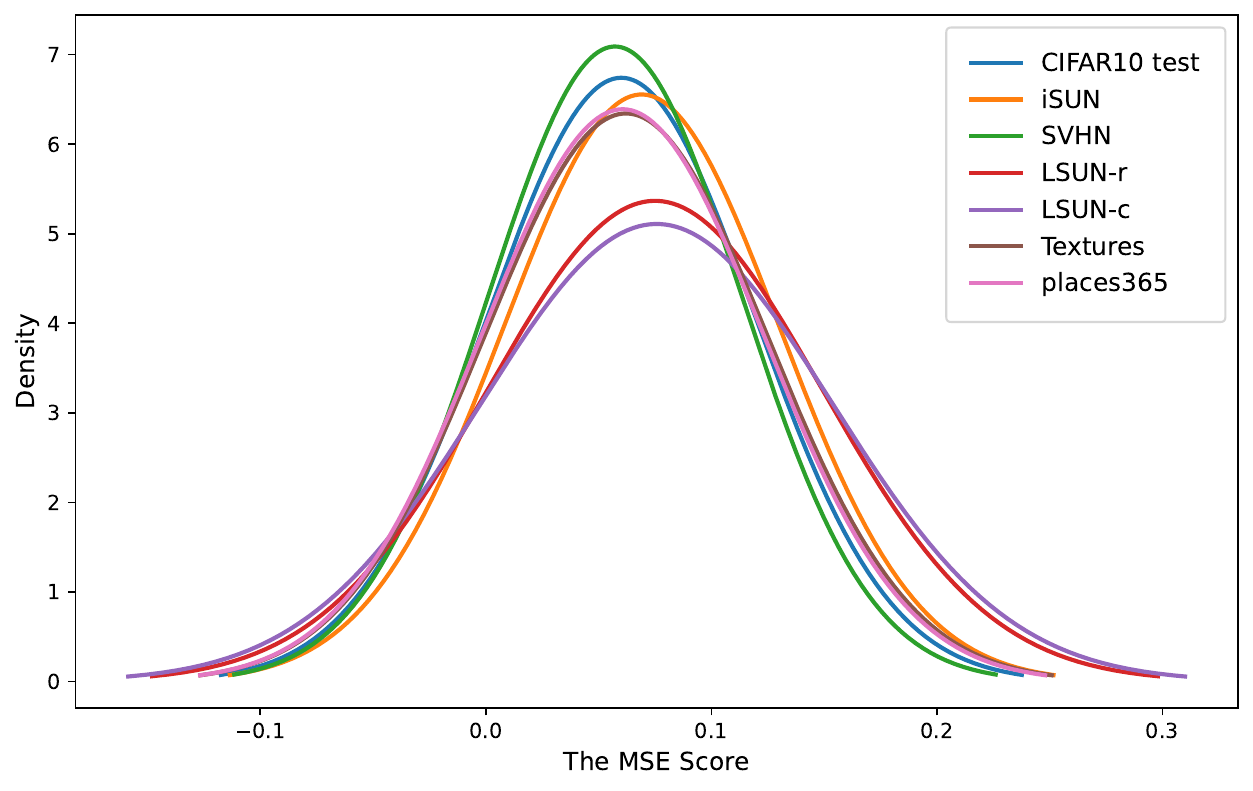}  % 设置宽度和高度
        \caption{Reconstruction Error Distribution of ID and OOD Samples for Pixel-level}
        \label{piexl}
    \end{minipage}
    \hfill
    \begin{minipage}{0.48\linewidth}
        \centering
        \includegraphics[width=\linewidth, height=4cm]{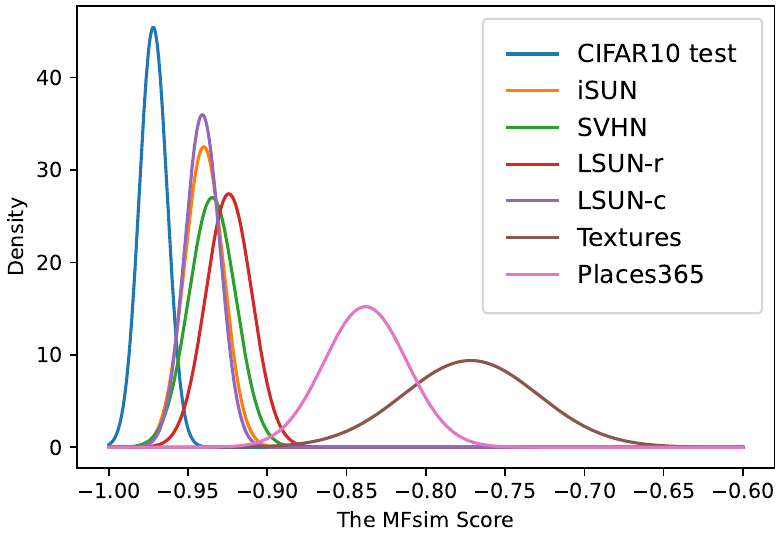}  % 设置宽度和高度
        \caption{Reconstruction Error Distribution of ID and OOD Samples for Feature-level}
        \label{feature}
    \end{minipage}
    \label{combined_distributions1111}
\end{figure}

We provide the distribution differences of the MSE score and MFsim score at two levels after training, with CIFAR-10 as ID dataset and other datasets as OOD; The results are shown in Figures \ref{piexl} and \ref{feature}.

It can be observed that at the pixel level(DDPM), the reconstruction error distributions of ID and OOD samples are very similar. The mixed MSE scores make it very hard to distinguish ID samples from OOD samples. However, at the feature level, the reconstruction score distribution of ID samples shows a clear distinction from that of OOD samples. The reason is that, our feature-level diffusion-based generative model makes the projected in-distribution latent space not only be compressed sufficiently to capture the exclusive characteristics of ID images, but also provide sufficient reconstruction power for the large-scale ID images of various categories. In other words, the pretrained encoder has inherent generalization capabilities, and the multi-layer features it extracts are more discriminative than the high-dimensional pixels of the images themselves.

\begin{table}[htbp]
  \centering
  \caption{The AUROC values compared to the other generative models using CIFAR-10 as the ID dataset.}
  \resizebox{\textwidth}{!}{
    \begin{tabular}{cl|c|c|c}
    \toprule
    \multicolumn{2}{c|}{Dataset} & \multicolumn{3}{c}{Method} \\
    \midrule
    \multicolumn{1}{c|}{ID} & \multicolumn{1}{c|}{OOD} & AE(+MFsim) & VAE(+MFsim) & Diffusion(+MFsim) \\
    \midrule
    \multicolumn{1}{c|}{\multirow{7}[2]{*}{CIFRA10}} & \multicolumn{1}{c|}{SVHN} & 57.68 & 83.96 & 98.89 \\
    \multicolumn{1}{c|}{} & \multicolumn{1}{c|}{LSUN} & 81.47 & 97.69 & 99.83 \\
    \multicolumn{1}{c|}{} & \multicolumn{1}{c|}{MNIST} & 95.85 & 99.98 & 99.99 \\
    \multicolumn{1}{c|}{} & \multicolumn{1}{c|}{FMNIST} & 79.61 & 98.69 & 99.99 \\
    \multicolumn{1}{c|}{} & \multicolumn{1}{c|}{KMNIST} & 90.51 & 99.96 & 99.99 \\
    \multicolumn{1}{c|}{} & \multicolumn{1}{c|}{Omniglot} & 81.50 & 97.69 & 99.99 \\
    \multicolumn{1}{c|}{} & \multicolumn{1}{c|}{NotMNIST} & 81.61 & 99.88 & 99.99 \\
    \midrule
        & \multicolumn{1}{c|}{average} & 81.18 & 96.84 & \textbf{99.81} \\
    \midrule
    Time & Num img/s (↑) & 1224.2 & 1179.4 & 999.3 \\
    \bottomrule
    \end{tabular}%
    }
  \label{tablle17}%
\end{table}%

\begin{table}[htbp]
  \centering
  \caption{The AUROC and FPR95 Values for Different Methods with ImageNet100 as ID Dataset and SUN/iNaturalist/Textures/Places365 as OOD Datasets.}
  \resizebox{\textwidth}{!}{
    \begin{tabular}{c|l|c|cc|cc|cc|cc|cc}
    \toprule
    ID & Method & \# img/s (↑) & \multicolumn{2}{c|}{SUN} & \multicolumn{2}{c|}{iNaturalist} & \multicolumn{2}{c|}{Textures} & \multicolumn{2}{c|}{Places365} & \multicolumn{2}{c}{Average} \\
    \cmidrule{4-13}
       &        &              & AUROC↑ & FPR95↓ & AUROC↑ & FPR95↓ & AUROC↑ & FPR95↓ & AUROC↑ & FPR95↓ & AUROC↑ & FPR95↓ \\
    \midrule
    \multirow{7}[4]{*}{ImageNet100} & MSP & 956.2 & 89.68 & 52.16 & 86.44 & 57.88 & 86.48 & 55.18 & 88.85 & 53.79 & 87.86 & 54.75 \\
       & EBO & 956.2 & 90.83 & 49.46 & 87.82 & 58.29 & 86.04 & 66.81 & 90.01 & 51.80 & 88.68 & 56.59 \\
       & DICE & 950.8 & 93.78 & 37.14 & \textbf{89.76} & \textbf{50.47} & 89.52 & 52.32 & 93.18 & 41.42 & 91.56 & 45.34 \\
       & ASH-S & 977.5 & 90.12 & 44.54 & 86.94 & 51.45 & 88.38 & 45.23 & 88.76 & 49.70 & 88.55 & 47.73 \\
    \cmidrule{2-13}        
       & ours(+MSE) & 954.9 & \textbf{98.81} & \textbf{5.91} & 86.59 & 71.88 & 97.68 & 10.76 & \textbf{98.79} & \textbf{5.73} & \textbf{95.47} & \textbf{23.57} \\
       & ours(+LR) & 313.4 & 93.98 & 16.51 & 51.39 & 87.99 & 76.02 & 47.54 & 94.95 & 16.25  & 79.09 & 42.07 \\
       & ours(+MFsim) & 954.9 & 98.49 & 8.41 & 84.63 & 77.86 & \textbf{98.08} & \textbf{9.98} & 98.37 & 8.75 & 94.89 & 26.25 \\
    \bottomrule
    \end{tabular}
  }
  \label{table11}
\end{table}
\subsection{ImageNet100 as ID Dataset}
In \textbf{Table~\ref{table11}}, our method using MSE outperforms the classification-based SOTA method DICE, achieving an improvement of 3.91\% in AUROC when ImageNet100 is used as the ID dataset and various datasets such as SUN, iNaturalist, Textures, and Places365 are used as OOD datasets. The significant improvements in performance metrics demonstrate that our generative-based approach can effectively model the in-distribution characteristics, leading to better OOD detection capabilities. This indicates that our proposed method is particularly suitable for more complex datasets like ImageNet100, where capturing detailed features is crucial for accurate OOD detection.

\section{Qualitative results.}
\begin{figure}[htbp]
    \centering
    % 第一幅图
    \begin{minipage}{\linewidth}
        \centering
        \includegraphics[width=1\linewidth]{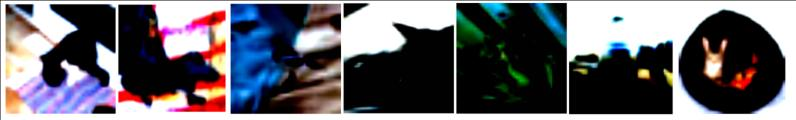}
        \caption{Examples of ID Samples Misclassified as OOD (Lacking Semantic Information).}
        \label{fig:image1}
    \end{minipage}

    \vspace{0.1cm} % 调整图之间的垂直距离

    % 第二幅图
    \begin{minipage}{\linewidth}
        \centering
        \includegraphics[width=1\linewidth]{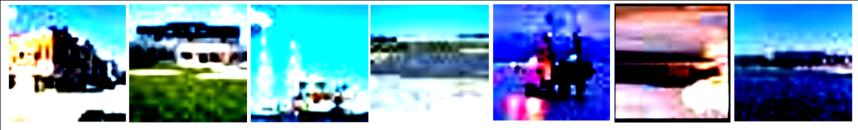}
        \caption{Examples of OOD Samples Misclassified as ID (Similar to ID Sample Categories).}
        \label{fig:image2}
    \end{minipage}

    \vspace{0.1cm} % 调整图之间的垂直距离

    % 第三幅图
    \begin{minipage}{\linewidth}
        \centering
        \includegraphics[width=1\linewidth]{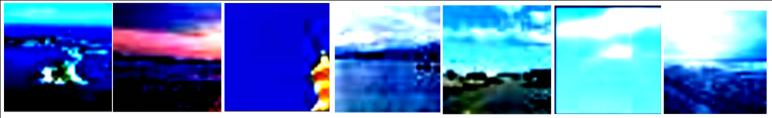}
        \caption{Examples of OOD Samples Misclassified as ID (Similar to ID Sample Colors).}
        \label{fig:image3}
    \end{minipage}
\end{figure}

We have included three types of failure cases Figures \ref{fig:image1}, \ref{fig:image2} and \ref{fig:image3} .
The first type, shown in Figure \ref{fig:image1}, represents ID samples misclassified as OOD. It can be observed that these misclassified samples often have significant shadows and lack semantic information, resulting in high reconstruction errors and being incorrectly classified as OOD samples.
The second type, shown in Figure \ref{fig:image2}, represents OOD samples misclassified as ID. It can be observed that these OOD samples have categories very similar to those of the ID samples (CIFAR-10), such as cars and ships, which are categories present in CIFAR-10.
The third type, shown in Figure \ref{fig:image3}, represents OOD samples with colors very similar to the ID samples, leading to their misclassification as ID.

% \section{Broader Impacts}
% \label{eeesss}
% \text{Positive Societal Impacts:} The proposed diffusion-based layer-wise semantic reconstruction method for unsupervised out-of-distribution (OOD) detection can significantly enhance the security and safety of machine learning systems. By effectively identifying OOD data, the system can prevent incorrect or potentially harmful decisions, making AI applications more reliable in critical areas such as healthcare, autonomous driving, and financial systems. This method increases the robustness of AI systems by ensuring they can handle unexpected inputs gracefully. This contributes to the overall stability and trustworthiness of AI deployments in various industries, thereby promoting wider acceptance and integration of AI technologies.
% \text{Negative Societal Impacts}: As with any advanced detection method, there is a risk that the technology could be misused. For instance, surveillance applications, it could be employed to monitor individuals without their consent, leading to privacy violations and ethical concerns.
% \section{ADDITIONAL RESULTS}

\newpage

\section*{NeurIPS Paper Checklist}

\begin{enumerate}

\item {\bf Claims}
    \item[] Question: Do the main claims made in the abstract and introduction accurately reflect the paper's contributions and scope?
    \item[] Answer: \answerYes{Yes}% Replace by \answerYes{}, \answerNo{}, or \answerNA{}.
    \item[] Justification: The main claims in the abstract and introduction accurately reflect our contributions. We propose a diffusion-based layer-wise semantic reconstruction method for unsupervised out-of-distribution (OOD) detection. Our method demonstrates superior performance in detecting OOD samples, as detailed in Section \ref{method} and Section \ref{Experiment}of our paper.

\item {\bf Limitations}
    \item[] Question: Does the paper discuss the limitations of the work performed by the authors?
    \item[] Answer: \answerYes{Yes} % Replace by \answerYes{}, \answerNo{}, or \answerNA{}.
    \item[] Justification: The limitations of our work are discussed in detail in Section \ref{hahha}
    % , where we outline the strong assumptions, scope of claims, factors influencing performance, and potential issues related to computational efficiency, privacy, and fairness.
    \item[] Guidelines:
    \begin{itemize}
        \item The answer NA means that the paper has no limitation while the answer No means that the paper has limitations, but those are not discussed in the paper.
        \item The authors are encouraged to create a separate "Limitations" section in their paper.
        , but those are not discussed in the paper.
        \item The authors are encouraged to create a separate "Limitations" section in their paper.
        \item The paper should point out any strong assumptions and how robust the results are to violations of these assumptions (e.g., independence assumptions, noiseless settings, model well-specification, asymptotic approximations only holding locally). The authors should reflect on how these assumptions might be violated in practice and what the implications would be.
        \item The authors should reflect on the scope of the claims made, e.g., if the approach was only tested on a few datasets or with a few runs. In general, empirical results often depend on implicit assumptions, which should be articulated.
        \item The authors should reflect on the factors that influence the performance of the approach. For example, a facial recognition algorithm may perform poorly when image resolution is low or images are taken in low lighting. Or a speech-to-text system might not be used reliably to provide closed captions for online lectures because it fails to handle technical jargon.
        \item The authors should discuss the computational efficiency of the proposed algorithms and how they scale with dataset size.
        \item If applicable, the authors should discuss possible limitations of their approach to address problems of privacy and fairness.
        \item While the authors might fear that complete honesty about limitations might be used by reviewers as grounds for rejection, a worse outcome might be that reviewers discover limitations that aren't acknowledged in the paper. The authors should use their best judgment and recognize that individual actions in favor of transparency play an important role in developing norms that preserve the integrity of the community. Reviewers will be specifically instructed to not penalize honesty concerning limitations.
    \end{itemize}

\item {\bf Theory Assumptions and Proofs}
    \item[] Question: For each theoretical result, does the paper provide the full set of assumptions and a complete (and correct) proof?
    \item[] Answer: \answerNA{\textcolor{gray}{N/A}} % Replace by \answerYes{}, \answerNo{}, or \answerNA{}.
    \item[] Justification: Our paper focuses on an experimental approach to out-of-distribution detection and does not include theoretical results. Therefore, this question is not applicable.

\item {\bf Experimental Result Reproducibility}
    \item[] Question: Does the paper fully disclose all the information needed to reproduce the main experimental results of the paper to the extent that it affects the main claims and/or conclusions of the paper (regardless of whether the code and data are provided or not)?
    \item[] Answer: \answerYes{Yes} % Replace by \answerYes{}, \answerNo{}, or \answerNA{}.
    \item[] Justification: Our paper fully discloses all necessary information to reproduce the main experimental results, including detailed descriptions of the experimental setup, datasets used, and evaluation metrics. This information is provided in Sections 4 of the paper.

\item {\bf Open access to data and code}
    \item[] Question: Does the paper provide open access to the data and code, with sufficient instructions to faithfully reproduce the main experimental results, as described in supplemental material?
    \item[] Answer: \answerYes{Yes} % Replace by \answerYes{}, \answerNo{}, or \answerNA{}.
    \item[] Justification: We have provided open access to our data and code, along with detailed instructions for reproducing the main experimental results. These resources are described in the supplemental material and can be accessed via the provided links.

\item {\bf Experimental Setting/Details}
    \item[] Question: Does the paper specify all the training and test details (e.g., data splits, hyperparameters, how they were chosen, type of optimizer, etc.) necessary to understand the results?
    \item[] Answer: \answerYes{Yes} % Replace by \answerYes{}, \answerNo{}, or \answerNA{}.
    \item[] Justification: Our paper specifies all necessary training and test details, including data splits, hyperparameters, and optimizer settings. These details are provided in Section 4.

\item {\bf Experiment Statistical Significance}
    \item[] Question: Does the paper report error bars suitably and correctly defined or other appropriate information about the statistical significance of the experiments?
    \item[] Answer: \answerYes{Yes} % Replace by \answerYes{}, \answerNo{}, or \answerNA{}.
    \item[] Justification: We have reported error bars for our main experimental results, calculated as the mean and standard deviation over three runs. Details on the calculation of error bars and the factors of variability considered (such as train/test split and random initialization) are provided in Section 4.

\item {\bf Experiments Compute Resources}
    \item[] Question: For each experiment, does the paper provide sufficient information on the computer resources (type of compute workers, memory, time of execution) needed to reproduce the experiments?
    \item[] Answer: \answerYes{Yes} % Replace by \answerYes{}, \answerNo{}, or \answerNA{}.
    \item[] Justification: The paper provides detailed information on the compute resources used for the experiments, including the type of compute workers (GPU), memory, and execution time. These details are specified in the experimental setup section \ref{Experiment}.

\item {\bf Code Of Ethics}
    \item[] Question: Does the research conducted in the paper conform, in every respect, with the NeurIPS Code of Ethics \url{https://neurips.cc/public/EthicsGuidelines}?
    \item[] Answer: \answerYes{Yes} % Replace by \answerYes{}, \answerNo{}, or \answerNA{}.
    \item[] Justification: We have thoroughly reviewed the NeurIPS Code of Ethics and confirm that our research conforms to these guidelines in every respect.

\item {\bf Broader Impacts}
    \item[] Question: Does the paper discuss both potential positive societal impacts and negative societal impacts of the work performed?
    \item[] Answer: \answerYes{Yes} % Replace by \answerYes{}, \answerNo{}, or \answerNA{}.
    \item[] Justification: We discuss the potential positive and negative societal impacts of our work in Section \ref{eeesss}. Specifically, we address how our method could improve unsupervised out-of-distribution detection, as well as the potential risks associated with misuse in surveillance applications.

\item {\bf Safeguards}
    \item[] Question: Does the paper describe safeguards that have been put in place for responsible release of data or models that have a high risk for misuse (e.g., pretrained language models, image generators, or scraped datasets)?
    \item[] Answer: \answerNA{N/A} % Replace by \answerYes{}, \answerNo{}, or \answerNA{}.
    \item[] Justification: Our paper does not involve the release of data or models that have a high risk for misuse. Therefore, this question is not applicable.

\item {\bf Licenses for existing assets}
    \item[] Question: Are the creators or original owners of assets (e.g., code, data, models), used in the paper, properly credited and are the license and terms of use explicitly mentioned and properly respected?
    \item[] Answer: \answerYes{Yes} % Replace by \answerYes{}, \answerNo{}, or \answerNA{}.
    \item[] Justification: We have properly credited the creators and original owners of the datasets and models used in our work. The licenses and terms of use are explicitly mentioned in Section 4 of our paper.

\item {\bf New Assets}
    \item[] Question: Are new assets introduced in the paper well documented and is the documentation provided alongside the assets?
    \item[] Answer: \answerYes{Yes} % Replace by \answerYes{}, \answerNo{}, or \answerNA{}.
    \item[] Justification: We have introduced new assets in the form of original code, and they are well documented. Detailed documentation is provided alongside the assets to ensure reproducibility and ease of use.

\item {\bf Crowdsourcing and Research with Human Subjects}
    \item[] Question: For crowdsourcing experiments and research with human subjects, does the paper include the full text of instructions given to participants and screenshots, if applicable, as well as details about compensation (if any)? 
    \item[] Answer: \answerNA{N/A} % Replace by \answerYes{}, \answerNo{}, or \answerNA{}.
    \item[] Justification: Our paper does not involve crowdsourcing nor research with human subjects. Therefore, this question is not applicable.

\item {\bf Institutional Review Board (IRB) Approvals or Equivalent for Research with Human Subjects}
    \item[] Question: Does the paper describe potential risks incurred by study participants, whether such risks were disclosed to the subjects, and whether Institutional Review Board (IRB) approvals (or an equivalent approval/review based on the requirements of your country or institution) were obtained?
    \item[] Answer: \answerNA{N/A} % Replace by \answerYes{}, \answerNo{}, or \answerNA{}.
    \item[] Justification: Our paper does not involve crowdsourcing nor research with human subjects. Therefore, this question is not applicable.

\end{enumerate}

\end{document}

%% file: macro.tex
\usepackage{bbm}

\usepackage{color}
\usepackage{CJKutf8}
\usepackage{svg}
\definecolor{DeltaColor}{rgb}{0.039,0.73,0.71}
\definecolor{SetaColor}{rgb}{0.867, 0.0235, 0.376}
\definecolor{SigmaColor}{rgb}{0.98,0.45,0.0}
\definecolor{RedColor}{rgb}{0.8,0,0}
\definecolor{AlphaColor}{rgb}{0,0,0.8}
\definecolor{BetaColor}{rgb}{0.8,0,0.8}
\definecolor{GammaColor}{rgb}{0.5,0,0.7}
\definecolor{EpsilonColor}{rgb}{0.353,0.725,0.906}
\definecolor{TauColor}{rgb}{0.423,0.235,0.192}
\definecolor{WtColor}{rgb}{0.235,0.470,0.470}

\usepackage{mfirstuc}